%% file: arxiv.tex
\documentclass[10pt,twocolumn,letterpaper]{article}

\usepackage{url}
\usepackage{tikz}
\usepackage{booktabs}
\usepackage{wrapfig}
\usepackage{multirow}
\usepackage{subcaption}
\usepackage{graphicx}
\usepackage{amsmath}
\usepackage{amssymb}

\usepackage[pagebackref,breaklinks,colorlinks]{hyperref}

\usepackage[capitalize]{cleveref}
\crefname{section}{Sec.}{Secs.}
\Crefname{section}{Section}{Sections}
\Crefname{table}{Table}{Tables}
\crefname{table}{Tab.}{Tabs.}

\usepackage{lipsum}

\newcommand\blfootnote[1]{%
  \begingroup
  \renewcommand\thefootnote{}\footnote{#1}%
  \addtocounter{footnote}{-1}%
  \endgroup
}
\begin{document}

%%%%%%%%% TITLE - PLEASE UPDATE
\title{Implicit Equivariance in Convolutional Networks}

\author{Naman Khetan$^*$,  Tushar Arora$^*$, Samee Ur Rehman and Deepak K. Gupta$^*$\\
Transmute AI Lab (Texmin Hub), IIT Dhanbad, India\\
Institution1 address\\
{\tt\small Naman.17je003197@am.iitism.ac.in, tushararora1410@gmail.com,}\\ 
{\tt\small s.u.rehman@icloud.com, deepak@transmute.ai}}
% For a paper whose authors are all at the same institution,
% omit the following lines up until the closing ``}''.
% Additional authors and addresses can be added with ``\and'',
% just like the second author.
% To save space, use either the email address or home page, not both
% \and
% Second Author\\
% Institution2\\
% First line of institution2 address\\
% {\tt\small secondauthor@i2.org}

\date{}
\maketitle

%%%%%%%%% ABSTRACT
\begin{abstract}

   Convolutional Neural Networks\blfootnote{$^*$Equal Contribution.\\ Code can be accessed at\\ \texttt{http://github.com/transmuteAI/implicit-networks}} (CNN) are inherently equivariant under translations, however, they do not have an equivalent embedded mechanism to handle other transformations such as rotations and change in scale. Several approaches exist that make CNNs equivariant under other transformation groups by design. Among these, steerable CNNs have been especially effective. However, these approaches require redesigning standard networks with filters mapped from combinations of predefined basis involving complex analytical functions. We experimentally demonstrate that these restrictions in the choice of basis can lead to model weights that are sub-optimal for the primary deep learning task (\emph{e.g.} classification). Moreover, such hard-baked explicit formulations make it difficult to design composite networks comprising heterogeneous feature groups. To circumvent such issues, we propose Implicitly Equivariant Networks (IEN) which induce equivariance in the different layers of a standard CNN model by optimizing a multi-objective loss function that combines the  primary loss with an equivariance loss term. Through experiments with VGG and ResNet models on Rot-MNIST , Rot-TinyImageNet, Scale-MNIST and STL-10 datasets, we show that IEN, even with its simple formulation, performs better than steerable networks. 
Also, IEN facilitates construction of heterogeneous filter groups allowing reduction in number of channels in CNNs by a factor of over 30\% while maintaining performance on par with baselines. 
The efficacy of IEN is further validated on the hard problem of visual object tracking. We show that IEN outperforms the state-of-the-art rotation equivariant tracking method while providing faster inference speed.

\end{abstract}

%%%%%%%%% BODY TEXT
\section{Introduction}
\label{sec:intro}
\input{latex/intro}

\section{Related Work}
\input{latex/related_work}

\section{Equivariance in CNNs}
\label{sec-exp-eq}
Equivariance refers to the property of a function to commute with the actions of any transformation group $G$ acting on its input as well as  output. Mathematically, any function $f: X \rightarrow Y$ is equivariant to the transformation group $G$, if
\begin{equation}
    f(\varphi_g^X(x)) = \varphi_g^Y(f(x)) \qquad \forall g \in G, x \in X,
\label{eq-abs-eq}
\end{equation}
where $\varphi_g^{X}$ and $\varphi_g^{Y}$ denote group actions in the respective spaces. Note that for $\varphi_g^Y = $id, equivariance simplifies to the special case of invariance.

CNN layers, involving transformation of feature maps $h$ through convolution with filters $\Psi$ are by architecture design equivariant under translations, \emph{i.e.}, $(\mathcal{T}_dh)*\Psi = \mathcal{T}_d(h * \Psi)$ where $\mathcal{T}_d$ is an action of the translation group $T \in (\mathbb{R}, +)$ that shifts the input by $d \in \mathbb{R}^2$. Apart from translation, there are often other transformations that occur in images, such as rotations, reflections and dilations. However, CNNs are not equivariant to such transformations by design. This limits the generalization of the model as patterns learnt for one orientation are not applicable to other orientations. To make  deep learning models robust with respect to these transformations, equivariant models are used.

\textbf{Steerable CNNs. }Steerable CNNs extend the notion of weight sharing to transformation groups beyond translations, thus making CNNs jointly equivariant to translations and other transformations. In the context of rotations, this implies performing convolutions with rotated versions of each filter. This weight sharing over  rotations leads to rotational equivariance in a manner analogous to how translational weight sharing in standard CNNs gives rise to translational equivariance. 

Steerable filters can be expressed as a linear combination of a fixed set of atomic basis functions $ \left \{ \psi_q  \right \}_{q = 1}^{Q }$ such that the filter can be steered directly based on transformation of the input. Since we present our discussions in this paper primarily in the context of rotations, we briefly discuss the formulation of steerable filters for rotations. For this case, Gaussian radial bases can be used to create rotationally steerable filters for an arbitrary angle $\theta$. Such a rotationally steerable filter $\Psi : \mathbb{R}^2 \rightarrow \mathbb{R}$ satisfies the following property for all angles $\theta \in (-\pi, \pi]$ and for angular expansion coefficient functions $\kappa_q $, 
\begin{equation}
\rho_{\theta} \Psi(x) = \sum_{q = 1}^{ Q } \kappa_q (\theta) \psi_q (x)
\end{equation}
where $\rho_{\theta}$ is a rotation operator. With the above steerable filter formulation, the response of different orientations can then be conveniently expressed in terms of the atomic basis functions, 
\begin{equation}
(f * \rho_{\theta} \Psi)(x) = \sum_{q = 1}^{ Q } \kappa_q (\theta) (f * \psi_q) (x).
\end{equation}

\section{Implicit Equivariance}
\label{ref-sec-exp-equ}
\input{latex/sec-ien}

\section{Experiments}

To demonstrate the efficacy of IEN, we conduct three sets of experiments and discuss related insights. Note that the goal of the experiments is not to improve over the state-of-the-art, rather to demonstrate that simple implicit formulations can also learn equivariance to the extent required for performing at least at par with conventional equivariant methods. First we evaluate the effectiveness of IEN for the task of classification. We focus on the transformation groups of rotations, reflections and scale, however, IEN can be applied to other transformations as well with minimal modifications. We further study the performance of Heterogeneous IEN and 
investigate if it can reduce inference cost without compromising on model performance. Finally, to evaluate IEN on more complex problems, we analyze its performance on the hard problem of visual object tracking in terms of inference speed and accuracy. Details related to all experiments including description of data and models, training procedure and associated hyperparameters is provided in the supplementary material.

\textbf{Baselines. } Since we aim
%{Before defining the baselines, we reiterate that IEN does not intend to achieve perfect equivariance in the model under a certain transformation. Rather, it aims} 
to identify the optimal amount of equivariance required to maximize the performance of the model on the primary task, we compare different models in terms of performance on the classification task as well as in terms of success rate and precision  for tracking. We compare our IEN models with steerable CNNs which  induce equivariance in the network by design. We employ the best performing variant of steerable CNNs based on the survey presented in \cite{weiler2019general}, and refer to this method itself as E2CNN. We perform all comparisons at the same inference budgets. As a secondary baseline, we also study the performance of regular CNN models of similar inference budgets trained with data augmentation. For scale variation, We compare the results with three baseline models, the SS-CNN \protect\cite{ghosh2019arxiv} and the two recent implementations of steerable networks, namely Deep Scale Space (DSS) \cite{worrall2019neurips} and scale equivariant steerable networks (SESN) \cite{Sosnovik2020Scale-Equivariant}.  For the tracking problem, our baseline is RE-SiamFCv2 \cite{gupta2021rotation} (referred as RE-SiamFC), a rotation-equivariant variant of SiamFC tracker \cite{bertinetto2016fully}.

\textbf{Learning equivariance under rotations and reflections. }We conduct experiments on Rot-MNIST \cite{worrall2017harmonic}, Rot-TIM (Rotation versions of TinyImageNet (TIM) dataset) and R2-TIM datasets (Rotation+Reflection versions  of TinyImageNet (TIM) dataset). % and the results follow below.

\textit{Rot-MNIST classification. }This dataset comprises 12000 and 50000 MNIST digits in the train and validation sets, respectively, and each digit is rotated by a random angle between 0 and 360 degrees. Additional details are provided in the supplementary material. For baseline, we implemented E2CNN with rotation groups comprising 4 equidistant orientations, thus $\Lambda=4$. Similarly, for IEN, we choose feature groups of size 4 in each layer of the network. Qualitative results related to this experiment are shown in Figure \ref{fig-mnist1} where we demonstrated that IEN learns the desired equivariance.

We implemented IEN with the same number of channels per layer as the E2CNN architecture. While our E2CNN implementation obtained an accuracy of 98.8, IEN achieved 98.6\% which is almost at par. Interestingly, the regular CNN model with channels equivalent to E2CNN and IEN, performed quite well on this dataset, with only a marginal drop in accuracy (98.4\%). This implies that even though CNN4-aug does not exhibit equivariance (as shown in  Figure \ref{fig-mnist1}), it is still sufficiently good for Rot-MNIST when trained with data augmentation. Nevertheless, Rot-MNIST is a relatively easy dataset, and we believe  that scores on this dataset should not be used as a measure to rank the models. We also briefly experimented with implementation of heterogeneous IEN models for this problem. We replaced 25\% of the feature groups of size 4 with 2 and 25\% with 1, thereby reducing the number of feature maps in every layer by 31\%. Even with this significance reduction in model size, no drop in model performance is observed. Details  are provided in the supplementary material.

\input{latex/table_rottim}
\textit{Rot-TIM classification. }Rot-TIM is the rotation variant of TinyImagenet (TIM) comprising 100 classes. Since Imagenet  is considered  a challenging dataset, we believe that Rot-TIM  is a suitable choice for analyzing equivariance of models under rotations. Tables \ref{table-resnet1} and \ref{table-vgg} present results obtained with variants of ResNet18 and VGG models, respectively. To design the various models, we first create E2CNN variants with the same number of parameters as standard VGG and ResNet18 architectures and then build regular CNN and IEN models that match the number of channels per layer of E2CNN. For VGG, equivariance loss is computed after every layer. For ResNet18 variants, we compute equivariance loss after every block. Within the block, $\beta_i=0$. We experiment with values of 0.01, 0.1 and 1.0 for $\beta_i$ in the IEN objective function and report the scores for best configurations here. Details related to the datasets and architectures follow in the supplementary material.

For almost all cases, IEN outperforms E2CNN for the same inference cost. For the ResNet18 experiment with R8 elements, IEN improves over E2CNN by a  margin of more than 3\% in accuracy. We also report scores with CNN models (same in size as E2CNN inference model) with and without data augmentation. Interestingly with the scaled up size, the regular CNN models perform fairly well. With data augmentation, the regular CNN models often outperform E2CNN. This typically happens when the equivariance error in the last layer of such networks is similar to that of IEN. 
%For example, equivariance error in the last layer is $\sim0.1$ for CNN4-aug as well as IEN on Rot-TIM dataset for the ResNet18 experiment. 
The level of equivariance at the last convolutional layer is the determining factor for model performance, and regular CNN models are often able to steer it accordingly (see the supplementary material for details). This implies that for certain cases, even without imposing the equivariance constraint, performance on the primary task can be maximized. However, since this cannot be known beforehand, implicit equivariance formulation should be a preferred choice.

\textit{R2-TIM classification. }This dataset is an extended version of Rot-TIM with random flips optionally applied on every sample of the dataset. IEN and CNN with augmentations outperform E2CNN on ResNet, see Table \ref{table-resnet1}. The score of CNN is slightly higher than IEN, which implies that imposing equivariance does not help in this case. An exception to the performance of IEN is the VGG experiment with R2-TIM dataset where E2CNN achieves the highest score (Table \ref{table-vgg}). With improved training procedure, we hope that IEN should match E2CNN scores for this experiment as well. %We believe that with improved training procedure, IEN should match E2CNN scores for this experiment as well. 

%\textit{R2-TIM classification. }This dataset is an extended version of Rot-TIM with random flips optionally applied on every sample of the dataset. From Table \ref{table-resnet1}, we see that IEN as well as CNN with augmentations outperform E2CNN on the ResNet experiment. Further, we see that the score of CNN model is slightly higher than IEN, which implies that imposing equivariance under reflections adversely affects the performance of the model. An exception to the performance of IEN is the VGG experiment with RR-TIM dataset where E2CNN achieves the highest score (Table \ref{table-vgg}). With improved training procedure, IEN should match E2CNN scores for this experiment. %We believe that with improved training procedure, IEN should match E2CNN scores for this experiment as well. 

\textbf{Equivariance with heterogeneous filter groups. }We explore here the potential of IEN in the context of improving inference speed. We employ heterogeneous groups in the model (denoted as Het-IEN) and evaluate the performance (results shown in Table \ref{table-het}). Our results show that Het-IEN implementations mostly outperform the baseline E2CNN scores while reducing the number of channels per layer of the network by more than 30\%. This translates into reduction of the filter blocks between every two layers by approximately a factor of 2, thereby significantly improving the inference speed. Detailed results are presented in Table \ref{table-het}

\textbf{Learning equivariance under change of scale. }For the generalizablity of IEN across different transformations, we also investigated its application for handling scale variations. 
% We compare the results with three baseline models including two recent implementations of steerable networks, namely Deep Scale Space (DSS) \cite{worrall2019neurips} and scale equivariant steerable networks (SESN) \cite{Sosnovik2020Scale-Equivariant}. 
We conducted experiments on Scale-MNIST and STL-10 datasets, and the results are described in 
Table \ref{table-scale}. On Scale-MNIST with images of size $28 \times 28$ pixels, SS-CNN and SESN obtained error scores of 1.95 and 1.76, respectively. In the similar setting, IEN obtained an average error score of 1.78 over three runs, performing at par with IEN. For images of size $56 \times 56$ , IEN achieves an error score of 1.33 compared to 1.42 and 1.57 obtained by SESN and DSS, respectively. This shows that IEN achieves performance similar to or better than the implicit formulation SESN, which, in this case is the equivalent explicit formulation of steerable networks for scale equivariance. We implement IEN similar to the vector formulation of SESN described in \cite{Sosnovik2020Scale-Equivariant}. For STL-10 dataset, IEN obtains a mean classification error of 10.11\% over three runs compared to 10.83\% and 11.28\% obtained for SESN and DSS, respectively. These results clearly demonstrate that the implicit learning of equivariance in IEN leads to improved classification performance even under variations of scale.

\input{latex/table_track}
\textbf{Rotation Equivariance in Object Tracking.} Table \ref{table-track} presents the results obtained with various models on Rot-OTB dataset \cite{gupta2021rotation}. All tracking algorithms presented here are variants of SiamFC \cite{bertinetto2016fully}, a popular algorithm for object tracking problems. As evaluation metrics, we use the commonly employed area under the curve scores for success and precision values \cite{wu2013online}. As  shown already in \cite{gupta2021rotation}, making SiamFC equivariant to rotations makes the tracker more robust to this transformation, see scores for RE-SiamFC. Details related to  setup of tracking experiments are  in the supplementary material.

Table \ref{table-track} shows the results obtained with the IEN implementation of SiamFC (referred as IEN-SiamFC). We show that IEN-SiamFC slightly outperforms RE-SiamFC for this task as well. Interestingly, IEN required very minimal modifications over the original SiamFC implementation, while RE-SiamFC required significant modifications to the network design. We further present tracking scores with a Het-IEN implementation that reduces number of channels per layer by 31\%. The results show that even with this significant drop in the number of channels, our Het-SiamFC model performs well for the object tracking task. This result clearly shows that with IEN implementations, it is possible to design trackers robust to rotations, while also providing good inference speed.

\textbf{Limitations. }
%IEN is simple to implement and is designed with the application/deployment point of view where budget at inference time is more important. Aligned with this, IEN shows significant gain in inference speed, however, it also comes at the cost of training additional parameters. 
IEN requires training of additional parameters. While our current choice of weighting terms used for equivariance loss works well for the chosen problems, it might not be the best choice for other problems. A rigorous study on this aspect is needed to understand the full potential of IEN. We choose mean-squared error as the measure of equivariance loss, however, limitations of this choice still need to be investigated. Training IEN becomes more complex when the weighting term for equivariance loss is chosen differently for each layer. We have only briefly investigated varying the weight of equivariance loss across different layers of the network, and the model performance is sensitive to it. Thus, for other problems, it might be of interest to vary this weighting term. 

\section{Conclusions}
In this paper, we have demonstrated that while restricting the basis choice for network weights to certain set of analytical functions can help the model to achieve perfect equivariance under certain transformations, it can be sub-optimal in performance on the primary deep learning task. To circumvent this issue, we have presented an implicit equivariance formulation, referred to as Implicit Equivariance Networks (IEN), and demonstrated through several numerical experiments that IEN can induce the required amount of equivariance needed in the network  to maximize the performance on the primary task. Our IEN strategy requires minimal modifications of the existing CNN models and can significantly reduce  inference cost compared to SOTA methods for equivariant transformations. Further implications of our research results are presented in Section \ref{sec:intro} of this paper.

{\small
\bibliographystyle{ieee_fullname}
\bibliography{arxiv}
}
%\newpage
%\newpage
\appendix
\section*{Appendices}
\section{Experiments:Additional Details}

\subsection{Datasets}
\textbf{Rot-MNIST. }
Rot-MNIST Dataset is a variation of the popular MNIST dataset containing handwritten digits. In Rot-MNIST, the digits are rotated by an angle generated uniformly between 0 and 2$\pi$ radians. %Therefore, in Rot-MNIST the factors of variation are the rotation angle and the factors of variation already contained in MNIST, such as handwriting style.
Therefore, in addition to the variations induced by the different handwriting styles for the digits 0 to 9, Rot-MNIST contains an additional source of variations, the rotation angle.
This dataset has been used as a benchmark in several previous papers \cite{worrall2017harmonic, weiler2018learning, weiler2019general}. It contains 12,000 images in the training dataset and 50,000 images in the validation dataset, where the size of each image is 28x28x1.
%Each input image has size of 28x28 and 1 channel and the label is one of the number between 0 to 9. While the data contains one randomly rotated variant of each sample in the train set, it contains several orientations of the same sample the validation set.

\textbf{Rot-TIM and R2-TIM. }
TIM stands for TinyImageNet Dataset which is a miniature version of the ImageNet Dataset \cite{deng2009imagenet}. TIM contains 200 classes. There are 500 images available for each class in the training set. On the other hand, the validation set contains 50 images for each class. All images have size [64x64x3]. 
We created 2 dataset variations of TIM, referred to as Rot-TIM and R2-TIM, each containing 100 classes that were randomly selected  from the 200 classes in the TinyImageNet Dataset. 
\newline
Rot-TIM is the Rotated variation of TIM where each image was rotated by an angle generated uniformly between 0 and 2$\pi$ radians. Therefore, Rot-TIM contains a total of 50,000 images for training across 100 classes where each class has 500 training images. The validation set for Rot-TIM, which contains 10,000 images, was created by sampling 2 rotated variants of each image from the original validation set of TIM. The rotation angles were once again sampled from a uniform distribution between 0 and 2$\pi$ radians. %For the validation set we sampled 2 rotated variants of each image in validation set of TIM generated uniformly between 0 and 2$\pi$ radians, making a total of 10,000 images in our validation set. 
\newline
R2-TIM is the Rotated \& Reflected variation of TIM. This dataset is the same as Rot-TIM apart from the fact that some images were flipped at random along the horizontal or vertical axis. 

%some images were flipped 
%it is similar to Rot-TIM with just a single change of having images being flipped along the horizontal or the vertical axis randomly along with rotations.

%Rot-TIM, we randomly selected 100 classes and rotated each image in the train dataset by a randomly selected angle between 0 to 2$\pi$ totalling to 40000 images in the train dataset. For validation dataset, we selected same set of 100 classes as chosen in the train dataset and generated 2 randomly rotated variants of each image totalling to 10,000 images in the validation dataset. RR-TIM stands for Rotated \& Reflected TIM. For RR-TIM, along with rotation, some images were also flipped along the horizontal or the vertical axis.

\textbf{Rot-OTB. }
Rot-OTB is a rotational variant of OTB100 dataset \cite{wu2013online} and was originally presented in \cite{gupta2021rotation} in the context of rotation equivariant object tracking. It contains 100 videos of the original OTB100 dataset, where the frames are rotated at a rate of 0.5 frames per second. In this paper, we use this dataset  for the purpose of validation of the object tracking task.

\textbf{GOT-10k. }
GOT-10k \cite{huang2019tpami} is a popular dataset for training object tracking models. %We use the training set for our various models trained in this paper.
We use this training set in order to train various object tracking models in this paper.
The training set comprises 9340 videos containing 840 classes of objects commonly found in daily life. Due to the diversity of objects available as well as the variations across the different challenges of object recognition, this dataset is popularly used in tracking.

\subsection{Evaluation Metrics}
\textbf{Validation Accuracy. }It denotes the percentage of samples that are correctly classified by the model from the total samples available in the validation set. The validation accuracy is therefore a measure of the ability of the learning algorithm to generalize to data not seen during training. 
%This is the commonly employed metric denoting how many samples from the validation set have been correctly classified out of the total samples. We employ this metric in the percentage notion. \dpk{reformulate this to make it more concrete.}

\textbf{Precision and Success. }Precision refers to the center location error and denotes the average Euclidean distance between the center location of the tracked targets and the manually labelled ground-truths \cite{wu2013online}. To understand the success score, we first need to understand overlap score. Overlap score denotes the ratio of intersection over union of the area of the predicted bounding box from a tracker and the manually labelled ground-truth. Success score denotes the average overlap score computed at various thresholds between 0 and 1. For more details, see \cite{wu2013online}.
\subsection{Training Hyperparameters}
In this section we provide details related to the  hyperparameters used during  training. We also discussed sensitivity to any new hyperparameter introduced in this paper.

\textbf{Rot-MNIST classification. }The E2CNN result reported in Table \ref{table-mnist} on the Rot-MNIST classification dataset was achieved using the same conditions as described in \cite{weiler2019general}. On the other hand, all results of standard CNN and IEN reported in Table \ref{table-mnist} are achieved using Adam optimizer.  One Cycle LR Policy \cite{DBLP:journals/corr/abs-1803-09820} was used to tune the Learning Rate. A cycle of 70 epochs was employed in which initially the Learning Rate was increased from 1e-5 to 5e-3 in the first one-fourth cycle and then decreased again to 1e-5 in the remaining part of the cycle.  After cycle completion, we continued training for 20 further epochs at a constant LR of 1e-5. Apart from the above mentioned set of hyperparameters, we tried several combinations of LR decay (exponential decay, one cycle policy with cycle of 40, 60 and 70 epochs), LR range (5e-3 to 1e-5 and 1e-2 to 1e-4) and weight decay (1e-5 and 1e-7) and reported the best results across these hyperparameter variations for each method in the table. Batch size of 64 was used for all experiments and $\beta_i$ = 0.01, 0.1, 0.5, 1.0, 10, 100, 1000 were experimented. Among these, $\beta_i$ = 1 exhibited highest performance which we have reported in the main paper.  %Table  \ref{table-mnist} also shows the extent of equivariance achieved by the different methods on each conv layer. 
% in We tried dpk{Mention all hyperparameters related to all the experiments of E2CNN, IEN - epochs, learning rate, weight decay}. describe which all $\alpha_i$ are 0 and which are non-zero. State details of which all $\alpha_i$ you have experimented.

\textbf{Rot-TIM and R2-TIM classifications. }For both IEN and E2CNN, the same set of hyperparameters were used as the ones used in Rot-MNIST classification experiments with just the batch size reduced to 32.
%yield highest accuracy with same hyperparameters as in the experiments of IEN in rot-MNIST experiment for both Resnet18 and VGG except batch size which is 32 in this case. E2CNN yeild highest accuracy when used one cycle policy for LR with cycle of 40 epochs and learning rate range 1e-2 to 1e-4 and optimized using batch size of 64 with Adam optimizer and weight decay of 1e-5. 
We tried $\beta_i$ = 0.1, 0.01 and 1 for all experiments of IEN. The best results are mentioned in the main paper while results for all values of $\beta_i$ are mentioned in Table \ref{table-rottim}.  In addition for both E2CNN and IEN  ResNet18 architectures  we interpolated the size of the image from 64 to 65 using bilinear interpolation so that after each convolution the spatial size comes out to be a perfect integer therefore allowing the equivariance to propagate to later layers. 
% \dpk{Mention all hyperparameter details for standard and adaptive IEN implementations.}

\textbf{Object tracking on Rot-OTB. }All training and inference conditions are chosen to be the same as those described in \cite{gupta2021rotation}. We experimented with $\beta_i = 0.01, 0.1 \text{ and } 1.0$. Among these, we found that the performance was best with $\beta_i = 0.1$. We reported this performance in the main paper. We believe that the performance on tracking should be more stable with respect to the choice of $\beta_i$ and we are exploring this as a part of our ongoing follow up research on the subject.

\subsection{Compute Resources}
All Rot-MNIST experiments have been completed using Google Colab Pro. For experiments on TIM dataset, we used a machine with 1 Nvidia GeoForce Titan X card. For tracking experiments, we use 1 RTX GPU with 32Gb memory.

\section{Additional Results}

\subsection{Results for Equivariance under heterogeneous filter groups}
Results for Equivariance under heterogeneous filter groups are shown in Table \ref{table-het}
\label{sec_app_het}
\input{latex/table_het}

\input{latex/table_mnist1.tex}
\subsection{Classification of MNIST digits}
The architecture for E2CNN was adopted from  \cite{weiler2019general}. Equivalent standard CNN and IEN architectures were implemented such that the number of inference parameters used by these models was equal to E2CNN. The models contain 6 convolutional layers. Full results for the experiments on Rot-MNIST dataset for E2CNN, IEN, and standard CNN as well as the  extent of equivariance achieved in each layer measured by the amount of equivariance loss in each layer is reported in Table \ref{table-mnist}. Also, the extent of equivariance observed in  these models is shown in Figure \ref{fig-mnist-equi} for all 6 \texttt{conv} layers. From the figure, it can be seen that  equivariance is achieved in each layer of IEN.
\input{latex/rotMNIST_exp}

\subsection{Results with Rot-TIM and R2-TIM}
\input{latex/Resnet18_exp.tex}
% An E2CNN model equivalent to ResNet18 and VGG was developed with the same layers and equal number of total trainable parameters. To maintain equal parameters count, a ratio was found in channels of the standard architecture and feature fields of E2CNN for different equivariance type. Full details about number of channels per layer in the base model and in E2CNN corresponding to equal trainable parameter count is provided in Table \ref{table-resnet-params} and \ref{table-vgg-params}.\\
\input{latex/table_resnet_params.tex}
\input{latex/table_vgg_params.tex}

% In addition, we developed the IEN architecture such that it exactly matches the architecture of E2CNN at inference time. %Full results for the experiments on Rot-TIM and RR-TIM dataset for E2CNN, standard CNN and IEN experiments with different values of $\beta_i$ and the extent of equivariance achieved in all \texttt{conv} blocks in case of ResNet18 and in 4 alternate conv layers in case of VGG (\texttt{conv}1, \texttt{conv}3, \texttt{conv}5, \texttt{conv}7) are reported in Table \ref{table-rottim}.
% Full results for the experiments on Rot-TIM and RR-TIM dataset for E2CNN, standard CNN and IEN experiments with different values of $\beta_i$ are shown in Table \ref{table-rottim}. The extent of equivariance achieved in all \texttt{conv} blocks in case of ResNet18 and in 4 alternate conv layers in case of VGG (\texttt{conv}1, \texttt{conv}3, \texttt{conv}5, \texttt{conv}7) are also reported in Table \ref{table-rottim}.
% Since the extent of equivariance is computed using the equivariance loss, lower values are better. The extent of equivariance observed in case of R8 in ResNet18 based E2CNN and standard CNN models is shown in Figure \ref{fig-resnet-e2cnn} and Figure \ref{fig-resnet-cnn8}, respectively. The corresponding results for IEN ($\beta_i$ = 0.01), IEN ($\beta_i$ = 0.1) are provided in Figure \ref{fig-resnet-een001}, Figure \ref{fig-resnet-een01} and Figure \ref{fig-resnet-een1}, respectively.

\subsection{Results of IEN under scale variations }
\label{app_scale_var}
Results related to performance of IEN under scale variations are shown in Table \ref{table-scale}.
\input{latex/sec_app_scale_var2.tex}

\input{latex/table_rottim_appendix2.tex}

\end{document}

%% file: latex/intro.tex
Over the last decade, state-of-the-art deep learning algorithms have continued to push the accuracy on computer vision tasks such as classification on challenging datasets  \cite{deng2009imagenet}. The translation equivariance property inherent to CNNs has been instrumental in enabling this performance. However, the vulnerability of CNNs to transformations such as rotation and scaling remains a challenge \cite{cohen2016group, weiler2018learning, Sosnovik2020Scale-Equivariant}, since standard CNNs do not learn features that are equivariant with respect to these transformations. 

One way to circumvent the above issue is to design networks invariant to such transformations. However, for deep learning problems, intermediate layers should not be invariant so that the relative pose of local features is preserved for the later layers \cite{cohen2016group, hinton2011auto}. Therefore, several approaches have been proposed to design networks that are \emph{equivariant} to certain transformation groups. A network is defined to be  equivariant if the output that it produces transforms in a predictable way on transformations of the input. Among the various methods that exist for constructing equivariant representations, steerable CNNs are particularly effective since variants of this approach have produced SOTA results for transformation groups of rotation, reflection and scale change on deep learning tasks of classification \cite{cohen2016group, cohen2016steerable}, segmentation \cite{weiler2018learning}, object tracking \cite{sosnovik2020scale, gupta2021rotation}, among others.

Steerable CNNs estimate filters for different orientations of a transformation group without the need for tensor resizing. The resultant architecture shares weights over the filter orientations, thereby improving generalization and reducing sample complexity. These methods essentially rely on the construction of filters from linear combinations of certain sets of basis functions. The choice of the basis can vary depending on the transformation group for which equivariance is to be induced. Examples include the use of circular harmonic functions 
\cite{worrall2017harmonic, weiler2018learning} 
and Hermite polynomials \cite{Sosnovik2020Scale-Equivariant} 
for rotation- and scale-equivariance, respectively.

Interestingly, while the basis can change across different constructions of steerable CNNs, the trained models are mostly similar to regular CNNs at inference time except that the filter weights differ. Based on this observation, we hypothesize that it should be theoretically possible to achieve absolute equivariance with regular CNN models as well - we need to find a model training procedure that biases the weights to converge towards portions of the optimization landscape that induce equivariance. However, as no simple mechanism exists that can force regular CNNs to learn equivariant representations, steerable CNNs, given their SOTA performance for rotation- and scale-equivariance on popular deep learning tasks \cite{weiler2019general, Sosnovik2020Scale-Equivariant, gupta2021rotation}, have thus far been preferred. 

It is important to note that while inducing equivariance in the network is a desirable property for certain deep learning problems, the primary goal is still to maximize its performance on tasks such as classification, regression and segmentation. Given this  goal, we posit that by restricting the weight parameterization space so that it can only be represented by combinations of certain basis sets, we are overly constraining the model. While such a strategy makes the CNN model equivariant to the desired transformation group, the performance on the actual learning task may be sub-optimal. Later in this paper, we validate this argument through numerical experiments. We show that for certain scenarios, regular CNN models outperform steerable CNNs given the same number of channels. Equivariance is only meant to help the model generalize better on a task in a data efficient manner, and we propose that the extent of equivariance required in a model could alternatively be discovered through the training process. This further implies that equivariance to a certain transformation group should be treated as an additional objective for which the CNN model should be trained for.

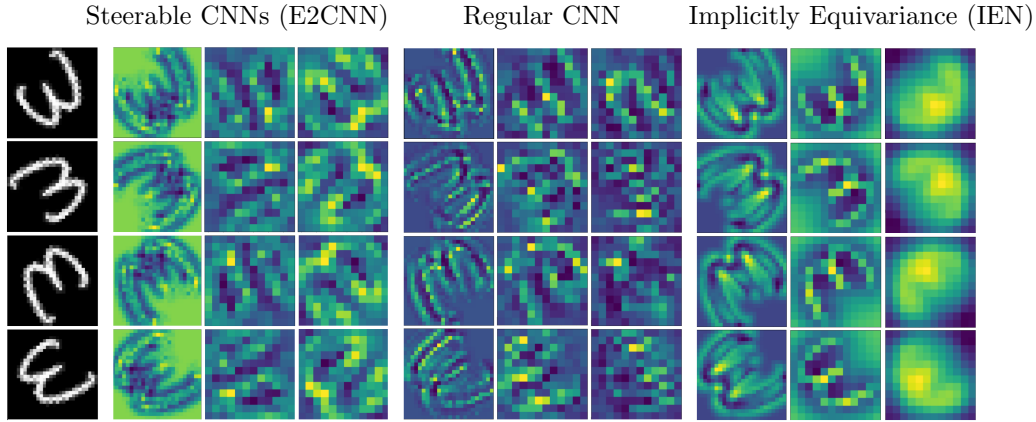
\begin{figure*}[t]
\centering
\begin{tikzpicture}
\node at (0,0) {\includegraphics[width=0.8\linewidth]{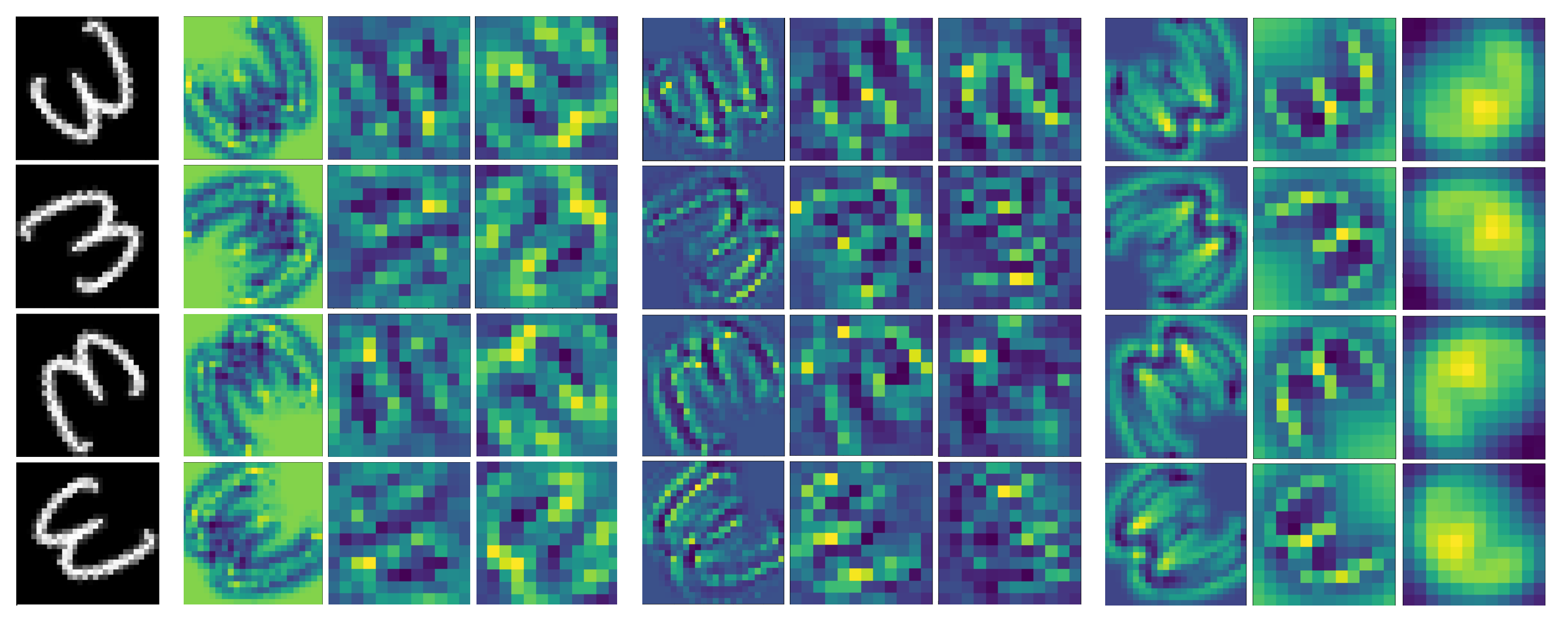}};
\node at (0.65, 2.9) {Regular CNN};
\node at (-3.4, 2.9) {Steerable CNNs (E2CNN)};
\node at (4.9, 2.9) {Implicitly Equivariance (IEN)};
\end{tikzpicture}
\caption{Feature maps obtained after first, third and fourth convolutions for 4 rotated versions of an MNIST digit with steerable networks (E2CNN), CNN and Implicitly Equivariant Network (IEN). }
\label{fig-mnist1}
\end{figure*}

Another limitation of steerable CNNs is that not only do they require redesigning  standard networks by incorporating sets of complex basis functions, but also, these basis functions tend to  vary for different transformation groups \cite{weiler2018learning, weiler2019general, Sosnovik2020Scale-Equivariant}. Significant effort would therefore be required to convert existing deep learning architectures into their group-equivariant counterparts. Furthermore, with such formulations, it is hard to design composite networks that are equivariant to multiple transformation groups \emph{e.g.}, rotations, reflections and change in scale. These networks also mostly work with homogeneous filter groups, where each filter group has the same group size. We hypothesize and later validate that not all filters require higher order discretizations. Therefore, it should be possible to achieve desired performance on given tasks with heterogeneous filter groups. An example would be to design convolutional layers with composite rotation groups of size 1, 2, 4 and 8 combined together. With reduced number of total channels, the inference speed of the model can be boosted. 

In this paper, we propose to achieve equivariance under a certain transformation group by adding constraints to the model training procedure. A better intuition of the idea can be obtained from Figure \ref{fig-mnist1}. For 4 different orientations of an MNIST digit, we show example feature maps for the first, third and fourth convolutional layers of E2CNN \cite{weiler2018learning, weiler2019general}, a regular CNN model and our Implicitly Equivariant Network (IEN). While the regular CNN fails to preserve equivariance, we see that E2CNN as well as IEN maintain equivariance under rotations. In terms of mean-squared error, the  extent of equivariance exhibited by E2CNN and IEN is on the order of $10^{-12}$ and $10^{-6}$, respectively. We observe that while IEN is less equivariant to rotations, it performs approximately at par with E2CNN in terms of classification accuracy on the Rot-MNIST dataset. We show later that for more complex datasets, IEN surpasses performance of E2CNN even with a lower level of equivariance. Based on the above observations, we formulate \emph{the implicit equivariance hypothesis}.

\textbf{ The Implicit Equivariance Hypothesis. }\emph{For a group-equivariant network where equivariance to a certain transformation group is hard-wired in the architecture by restricting the choice of filters to a certain subset (e.g., circular harmonics for rotation group), the performance of the trained model is not always optimal and standard CNN architectures when trained with additional constraints may be able to achieve better accuracy at same or lower inference cost than the former.}
%\textbf{ The Implicit Equivariance Hypothesis. }\emph{For a group-equivariant network where equivariance to a certain transformation group is explicitly hardwired in the network architecture by restricting the choice of filters to a certain subset (e.g., circular harmonic functions for rotation group), there exists a family of CNN architectures which, when trained with additional soft constraints, become sufficiently equivariant with accuracy at par or above the former at the same or even lower inference costs.}

Based on the hypothesis above, we transform the regular CNN model into a multi-objective formulation where an additional  equivariance loss term is optimized together with the primary loss component of the model. We refer to this approach as Implicitly Equivariant Network (IEN) in the rest of the paper and a more formal problem description is provided in \mbox{Section \ref{ref-sec-exp-equ}}. IEN makes existing CNN models robust to desired transformations with minimal modifications to the network. 

\subsection*{Contributions}
% \vspace{-1em}
\begin{itemize}
\itemsep-0.1em 
    \item We demonstrate that by implicitly constraining regular CNNs to learn equivariance during training, it is possible to achieve better performance on deep learning tasks compared to steerable CNNs. 
    \item IEN architectures perform at par with steerable CNNs while reducing the number of channels per layer by more than 30\%.  This reduction happens because IEN facilitates heterogeneous combinations of transformation groups of different orders. 
    \item In contrast to the difficulty of designing steerable CNNs that are equivariant under multiple transformation groups, we show that it is relatively simple to design IEN architectures that are  equivariant  under multiple transformations groups. 
    \item We show that IEN achieves performance at par with steerable networks while providing better inference speed over the latter by applying it on the problem of visual object tracking. 
    \item Our IEN formulation treats equivariance to a transformation group as an additional objective to be optimized. This ensures that absolute equivariance is not explicitly hardwired in the model. Rather, it is induced only to the extent that model performance on the primary task is maximized.    
\end{itemize}

\subsection*{Implications }With the concept of implicit equivariance outlined above, we hope to have paved the groundwork for easy and efficient implementation of equivariance in existing CNN architectures. We also hope that this research promotes further development on the following aspects.

\textit{Learning equivariance for black-box transformations. }With the simplicity of IEN, future CNN architectures can be made robust  against transformations for which the choice of analytical basis functions is not straightforward (\emph{e.g.}, occlusion).

\textit{Designing highly robust models for open-world scenarios. }Since IEN can combine equivariance to multiple transformation groups easily, advanced CNN architectures can be designed that are equivariant to a large set of transformations in a unified manner, thus making them more robust with respect to open-world problems. 

\textit{Easy integration with existing models. }Since IEN requires  minimal modifications to existing CNN architectures it can be easily integrated with existing CNN models. 

\textit{Improved inference speed. }Due to weight sharing, existing steerable CNNs use relatively larger number of channels per layer. This limits their use in problems where inference speed is crucial. Our heterogeneous IEN  formulations alleviate this issue, thereby increasing their scope of applicability.

%% file: latex/related_work.tex
% Robustness of computer vision algorithms with respect to changes of orientation, scale and perturbation has become an increasingly important research topic. In particular, the brittleness of deep learning methods with respect to slight changes in  orientation has also  gained attention \cite{DBLP:journals/corr/abs-1712-02779}. %It is becoming clear that the same statistical artifacts and correlations that the algorithms are using to become more accurate are the ones that make the learning engines prone to break in case of even slight demands of out of distribution generalization. 

%To address some of these challenges, research is being conducted to construct learning algorithms that give rise to equivariant representations. 
A commonly utilized technique to tackle variations of orientations and scale in computer vision algorithms is data augmentation.
This involves supplying transformed copies of the input data to the model which primarily helps in  learning invariance. However, this strategy can learn equivariance as well to a limited extent \cite{krizhevsky2017imagenet, lenc2015understanding, laptev2016tipooling}. Methods such as equivariant Boltzmann machines \cite{sohn2012learning} and equivariant descriptors \cite{6247909}  have  been proposed to construct learning algorithms that give rise to equivariant representations. A different set of algorithms focus on inducing equivariance in the model by design.  Collectively referred as group-equivariant networks, these involve reparameterization and obtain the filter weights by employing a set of atomic bases or by projecting to a different space.

A large body of work exists on designing group-equivariant convolutions. These include discrete roto-translations in 2D \cite{cohen2016group, weiler2018learning, bekkers2018miccai} and  \cite{worrall2018eccv}, continuous rot-translations in 2D \cite{worrall2017harmonic} and 3D \cite{weiler2018neurips, kondor2018neurips}, in-plane reflections \cite{weiler2018neurips, weiler2019general}, continuous rotations on the sphere \cite{esteves2018eccv}, and equivariance to discrete scales \cite{Sosnovik2020Scale-Equivariant}, among others. 
The GCNN approach \cite{cohen2016group} effectively learns equivariant representations but scales linearly in the number of orientations which prohibits its applicability on large-scale problems. This issue is overcome by steerable CNNs that generate equivariant  representation using composition of elementary features or atomic filters \cite{cohen2016steerable}. Subsequent improvements on these architectures have also been made recently \cite{weiler2018learning, weiler2019general}.
%Through the choice of appropriate basis, these methods transform the filter groups such that desired equivariance can be achieved. 
Among these, steerable CNNs have been most effective \cite{weiler2019general}. The idea of using steerable bases is not new and was proposed in earlier research \cite{freeman1991tpami, teo1997cvpr}. Our idea of IEN is inspired by the design of steerable CNNs presented in \cite{weiler2018learning}.

% Other approaches involving filter rotations such as \cite{Marcos_2016} and \cite{Marcos_2017} have been suggested but these depend on bicubic interpolations that introduce inaccuracies. Group-convolutional scattering transforms that depend on a fixed rather than learnt set of filters have also been proposed to learn equivariance \cite{6619007}. Methods that significantly alter the underlying learning architecture by replacing the scalar-output feature detectors based CNNs with transforming auto-encoders that learn features in the form of whole vectors of instantiation parameters have been proposed to accurately represent pose \cite{10.5555/2029556.2029562}. This has lead to the development of Capsule Networks to introduce robustness with respect to changes in orientation and/or viewpoint \cite{sabour2017dynamic, 46653}. 

Recently, an approach to learn the rotational versions of the filter through an additional equivariance loss has been presented \cite{pmlr-v97-diaconu19a}. However, it focused on learning to transform the filters without compromising the equivariance property of the model. This helps to tackle the adverse effects of \emph{post hoc} discretization suffered by  steerable CNNs \cite{weiler2018learning}. In contrast to this, our work focuses on learning the right balance between making the model equivariant and optimizing its performance on the primary deep learning task. A few approaches exist that use regularization to maintain consistency at the final output \cite{sajjadi2016neurips, berthelot2019neurips}, however, unlike IEN, these are limited to achieving invariance.
The approach of \cite{benton2020neurips} is closest to IEN since this strategy enforces equivariance at the end of the network through data augmentation. However, this approach focuses on learning the distribution of input and does not include an additional loss term to induce equivariance in the network, rather it relies on a combination of forward and inverse transforms of the input data and the output to achieve equivariance. Another recent work \cite{Castro2020SoftRE} has shown that through an added regularisation term soft equivariance could be achieved for classification task, however a robust and generic pipleline as well as application on more complex computer vision problems is still missing in the existing literature.

%% file: latex/sec-ien.tex
\textbf{General description. }\emph{Implicit equivariance} refers to the process of inducing equivariance in the model during training by optimizing it with respect to an additional loss term. We add an equivariance loss term to the objective function that  promotes exploiting parts of the weight space for which the network has relatively improved equivariance under desired transformations.  Unlike other  equivariance approaches \cite{weiler2018learning, weiler2019general, Sosnovik2020Scale-Equivariant}, the model is not forced to learn absolute equivariance. Rather, the implicit equivariance formulation achieves equivariance only to the extent that model performance on the primary task, \emph{e.g.} classification, is maximized.

%To induce equivariance in the model, we introduce \emph{feature groups}.
%Each feature group corresponds to a set of feature maps that undergo a pooling operation, referred to as \emph{group-pooling} or $\mathcal{G}(\cdot)$, that takes pixel-wise maximum values across all the feature maps within the feature group  to compute a pooled response.
%
%The network utilizes the presence of multiple feature maps within each group to learn the desired equivariance. This also implies that with only one feature map per group, the network tends to learn invariance.

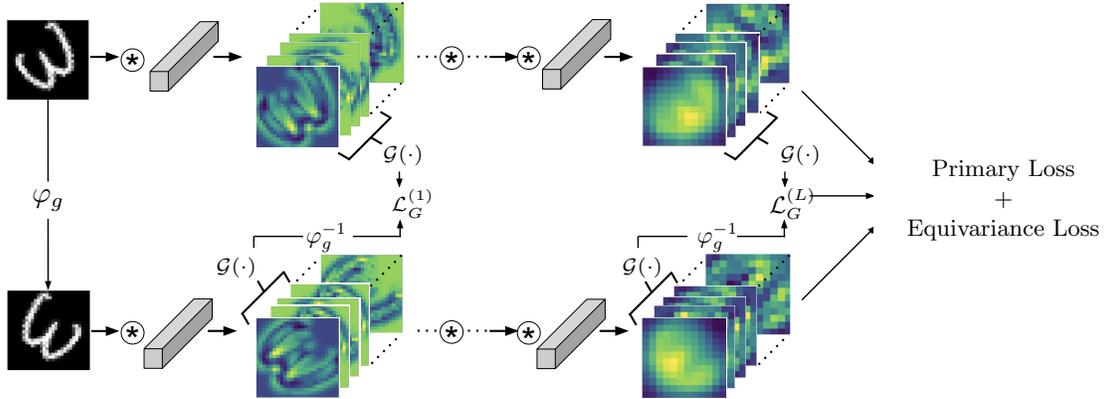
\begin{figure*}
% \caption {Schematic representation showing the working of Implicitly Equivariant Networks. Here $\varphi_g$, $\varphi_^{-1}_g$ and $\mathcal{G}(\cdot)$ denote rotation, rotation in opposite direction and group-pooling, respectively}
    \centering
\begin{tikzpicture}
%\node[draw] at (-5, -2)  (c)     {A}; 
\node[inner sep=0pt] (a1) at (0,0)
        {\includegraphics[width=.7\textwidth, trim =0 0 3.8cm 0, clip=true]{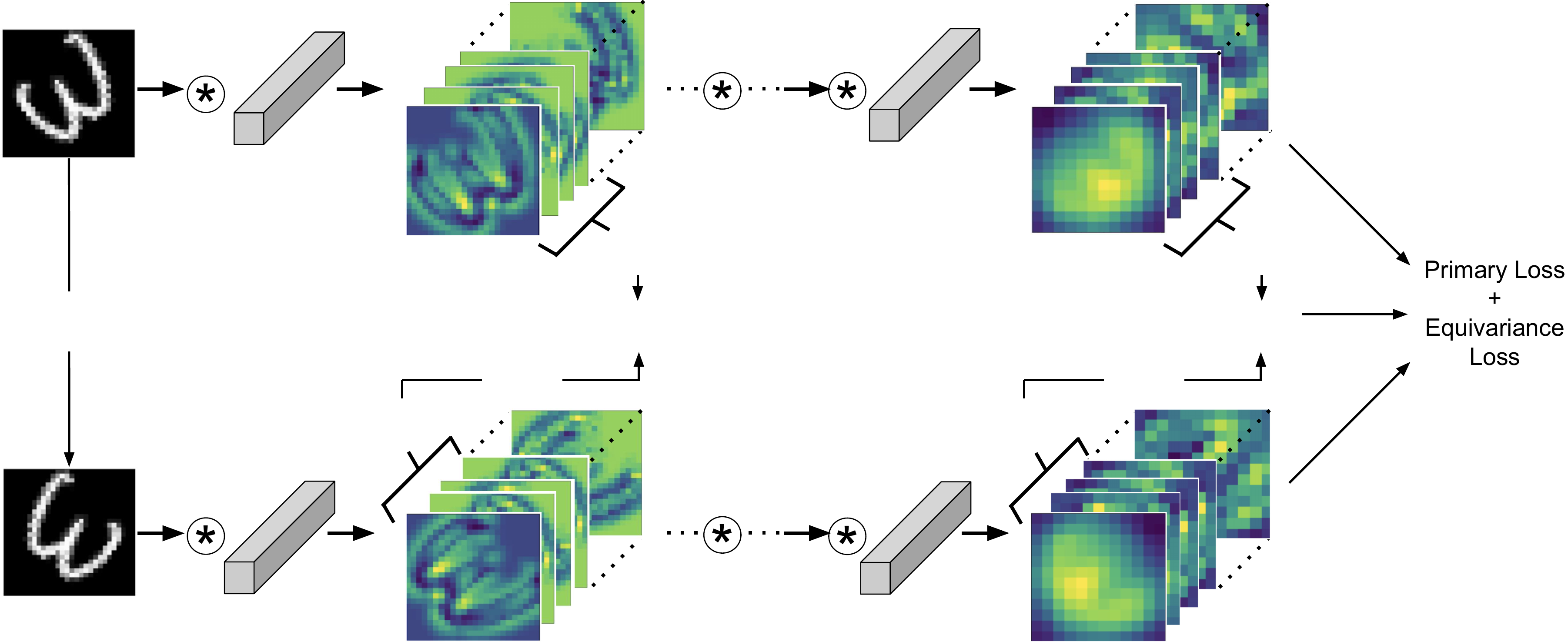}};
\node[scale = 1.1, color = black] at (-5.2, 0.0)  (a)     {$\varphi_g$};
\node[scale = 1.1, color = black] at (-2.7, -0.9)  (b)     {\scriptsize $\mathcal{G}(\cdot)$};
\node[scale = 1.1, color = black] at (2.7, -0.9)  (b)     {\scriptsize $\mathcal{G}(\cdot)$};
\node[scale = 1.1, color = black] at (-0.48, 0.6)  (c)     {\scriptsize $\mathcal{G}(\cdot)$};
\node[scale = 1.1, color = black] at (4.8, 0.6)  (c)     {\scriptsize $\mathcal{G}(\cdot)$};
\node[scale = 1.1, color = black] at (-0.35, -0.05)  (c)     {\scriptsize $\mathcal{L}_G^{(1)}$};
\node[scale = 1, color = black] at (4.7, -0.05)  (d)     {\small $\mathcal{L}_G^{(L)}$};
\node[scale = 0.9, color = black] at (3.7, -0.5)  (a)     {$\varphi^{-1}_g$};
\node[scale = 0.9, color = black] at (-1.5, -0.5)  (a)     {$\varphi^{-1}_g$};
\node at (7.5, 0.4){\small Primary Loss};
\node at (7.5, 0.0){\small +};
\node at (7.5, -0.4){\small Equivariance Loss};
\end{tikzpicture}
\caption 
{Schematic representation showing the working of Implicitly Equivariant Networks. 
Here {$\varphi_g$},
 {$\varphi_g^{-1}$} and
{$\mathcal{G}(\cdot)$} denote
rotation, rotation in opposite direction and group-pooling, respectively.}
\label{fig-schem} 
\end{figure*}

Figure \ref{fig-schem} shows a schematic representation of the implicit equivariance concept. The network takes an input image and its transformed versions and  computes the primary loss $\mathcal{L}_{\tau}$, \emph{e.g.} loss corresponding to classification error, for each of these inputs. Additional loss terms $\mathcal{L}_G$ are computed from different parts of the network which are combined together to form the equivariance loss. We present the discussion primarily for rotation equivariance, however, IEN can be formulated for other transformations as well. We show this through using IEN for scale equivariance as well.

To simplify the implementation of the equivariance loss, we introduce the concept of a feature group. Each feature group corresponds to a set of feature maps that undergo a pooling operation, referred to as \emph{group-pooling} or $\mathcal{G}(\cdot)$, that takes pixel-wise maximum values across all the feature maps within the feature group  to compute a pooled response. 
The equivariance loss computed at the $i^{\text{th}}$ layer corresponds to the pixel-wise error between the aligned versions of feature groups of the input and feature groups of the transformed versions. Therefore, the network utilizes the presence of multiple feature maps within each group to learn the desired equivariance. 
The max-pooling operation can introduce a small amount of invariance, but we use it as computing the equivariance loss over pooled feature maps is easier to implement than using the original feature maps directly. 

\textbf{Mathematical formulation. }We first provide a general mathematical formulation for the training of a standard CNN model for a certain deep learning task $\tau$.
We define the CNN mapping as $\mathcal{F}$ with weights $\mathbf{W}$ and a set of hidden channels $\mathbf{H}$. Further let $\mathbf{h}_i \in \mathbf{H}$ denote the set of channels at the $i^{\text{th}}$ layer of the network. Let $\mathcal{L}_{\tau}$ denote the loss function used to compute the performance for task $\tau$ during training. Based on this description, the mathematical statement describing the optimization of the CNN model is

\begin{equation}
\min_{\mathbf{W}} \mathcal{L}_{\tau}(\mathcal{F}( \mathbf{H(\mathbf{W}); \mathbf{x}}), \mathbf{y}), 
\label{eq_opt_problem}
\end{equation}

where $\{\mathbf{x}, \mathbf{y}\} \in \mathcal{D}$ are data samples used to train the network. 

Typically, group-equivariant models \cite{weiler2018learning, Sosnovik2020Scale-Equivariant} induce equivariance by computing the feature maps for every hidden layer of the network for $\Lambda$ different orientations. Intuitively, the $\Lambda$ orientations attempt to approximate the continuous transformation field, and the extent of equivariance improves with larger values of $\Lambda$. Inspired by their idea of filter groups, we split the feature maps $\mathbf{h}_i$ of the $i^{\text{th}}$ layer of $\mathcal{F}$ into groups of $\Lambda$. Based on this notion, we have $n(\mathbf{h}_i)/\Lambda$ feature groups for the $i^{\text{th}}$ layer of the network. Note that unlike the group-equivariant networks which share weights across sets of $\Lambda$ filters, no weight sharing is involved in IEN.  

\emph{Intermediate group-pooling. }In IEN, feature maps within every feature group are trained to collectively learn information for $\Lambda$ orientations. To extract the final outcome from the convolutional layers, a group-pooling layer is employed that computes  pixel-wise maximum value across all the orientations to construct the output feature map. 
%Group-pooling does not preserve equivariance, and part of the network that follows it becomes invariant under transformations of the input. Hence, for tasks such as classification, group-equivariant CNNs commonly employ it only after the last convolutional block, beyond which equivariance is no longer desired. See \cite{weiler2019general} for more details.
In addition to using group-pooling after the last convolutional layer of $\mathcal{F}$, IEN employs additional group-pooling at intermediate layers of the network where equivariance loss is computed.  These operations are performed on the hidden feature groups, $\mathbf{h}_{ij}$, where $i$ and $j$ are used as indices for the hidden layer of the network and the feature group, respectively. We add a group pooling operation $\mathcal{G}(\cdot)$ on every feature group, such that any resultant feature map $\hat{h}_i \in \hat{\mathbf{h}}_i$ is obtained by pooling over the $\Lambda$ orientations $\mathbf{h}_{ij}$, denoted as $\hat{\mathbf{h}}_i = \mathcal{G}(\mathbf{h}_{ij})$. 

Figure \ref{fig-schem} shows how feature maps pooled over the orientations are used to compute equivariance loss from different parts of the network. Note that while the equivariance loss in the intermediate layers is computed using $\hat{\mathbf{h}}_i$, the input to the next layer $\mathbf{h}_{ij}$ contains all the original feature maps. This means that the next layers are not impacted by this max-pooling operation and our intermediate group-pooling operations \emph{do not detrimentally affect} the equivariance in  later parts of the network.

\textit{Equivariance loss. }This loss term provides a measure of non-equivariance that exists in a model for any group action $\varphi_g$ from the transformations $G$. We define our equivariance loss measure across the network based on the absolute equivariance condition stated in Eq. \ref{eq-abs-eq} and express it in terms of mean-squared error as
\begin{equation}
    \mathcal{L}_G(x) = \sum_{g \in G}\|f(\varphi_g^X(x)) - \varphi_g^Y(f(x))\|_2^2 \qquad \forall x \in X.
    \label{eq-eqloss1}
\end{equation}
Here, $\mathcal{L}_G$ denotes equivariance loss computed over all group actions $\varphi_g$ from $G$. The term $f$ represents a subset of the IEN model $\mathcal{F}$, and typically comprises one or more convolutional layers, each followed by a nonlinear activation and a batch-normalization module. Based on the notion of feature groups defined earlier, equivariance loss for transformations $G$ at the $i^{\text{th}}$ layer can be stated as
\begin{equation}
\begin{split}
    \mathcal{L}_G^{(i)}(x_p, x_q) = \sum_{g \in G}\| \hat{\mathbf{h}}_i(x_p) - \varphi_g^Y(\hat{\mathbf{h}}_i(x_q))\|^2 \\ \text{s.t.} \enskip x_p = \varphi_g^X(x_q).
    \label{eq-eqloss2}
    \end{split}
\end{equation}
Here, $x_p$ and $x_q$ are two different inputs from the input set $\mathbf{x}$, and as stated in the equality constraint, one is the transformed version of other. 
%For ease of implementation, $x$ and $\varphi^X_g(x)$ are treated as two different input samples passed to the subnetwork $f$ as part of the same batch iteration step.
Based on Eq. \ref{eq-eqloss2}, the equivariance error across all $L$ layers of the network for the input set $\mathbf{x}$ is
\begin{equation}
\begin{split}
    \mathcal{L}_G(\mathbf{x}) = \sum_{i=1}^L\sum_{g \in G}\| \hat{\mathbf{h}}_i(x_p) - \varphi_g^Y(\hat{\mathbf{h}}_i(x_q))\|^2 \cdot \beta_i \\ \forall \enskip \{x_p, x_q\} \in \mathbf{x} \enskip \text{s.t.} \enskip x_p = \varphi_g^X(x_q).
    \label{eq-eqloss3}
    \end{split}
\end{equation}
% \begin{equation}
% \begin{split}
%     \mathcal{L}_G(\mathbf{x}) = \sum_{i=1}^L\sum_{g \in G}\| \hat{\mathbf{h}}_i(x_p) - \varphi_g^Y(\hat{\mathbf{h}}_i(x_q))\|^2 \cdot \beta_i \\ \forall \enskip \{x_p, x_q\} \in \mathbf{x} \enskip \text{s.t.} \enskip x_p = \varphi_g^X(x_q).
%     \label{eq-eqloss3}
%     \end{split}
% \end{equation}
The weighting term  $\beta_i$ provides control over  the amount of equivariance imposed in different parts of the network. For layers where equivariance is to be implicitly enforced, $\beta_i > 0$, else $\beta_i = 0$.

\textit{Multi-objective optimization. }Our IEN architectures are trained for multi-objective loss that combines primary loss $\mathcal{L}_{\tau}$ with the equivariance loss term. The IEN optimization problem is then stated as
\begin{equation}
\min_{\mathbf{W}}  \mathcal{L}_{\tau}(\mathcal{F}( \mathbf{H(\mathbf{W}); \mathbf{x}}), \mathbf{y}) + \sum_{G\in \mathcal{S}}\alpha_{G} \mathcal{L}_{G}(\mathbf{x}), 
\label{eq_opt_problem}
\end{equation}
where $\mathcal{S}$ denotes the set of transformations for which equivariance is to be learnt and $\alpha_{\mathcal{S}}$ is the corresponding weighting term. 

\textbf{Heterogeneous IEN. } An advantage of IEN  is that filter groups of different sizes can be combined together. For $N$ filter groups in steerable CNNs \cite{weiler2018learning, weiler2019general} as well as the $N$ feature groups in the IEN implementation  above, there would be $N\cdot\Lambda$ feature maps in the respective layer of the network. This number scales linearly in $N$ and $\Lambda$. Het-IEN relies on the assumption that not all features  need cyclic groups of size $\Lambda$. For example, for $\Lambda=8$, simpler features could be fully represented with lower order cyclic groups as well, for example 4, 2, or even 1. We refer to such combinations as heterogeneous feature groups. For example, by expressing $N$ feature groups of $\Lambda=8$ with $0.5N$ groups of 8 orientations, $0.25N$ groups of 4 orientations, $0.125N$ groups of 2 orientations and $0.125N$ groups of 1 orientation, the number of feature maps in the respective layer can be reduced by 33\%, which results in significant boost in inference speed.

%% file: latex/table_rottim.tex
\begin{table}
% \begin{minipage}{0.49\textwidth}
 \caption{Performance scores for ResNet18 and its equivariant versions on Rotation (Rot-TIM) and Rotation+Reflection (R2-TIM) versions of TinyImageNet (TIM) dataset. Here, CNN4-aug and CNN8-aug denote regular CNN models similar in architecture to those of R4 and R8, respectively, at inference phase. R4  and  R8  denote  equivariance  to  4  and  8  equidistant orientations, respectively, and R4R denotes equivariance to 4 equidistant rotations and reflections.}
  \label{table-resnet1}
  \centering
  \begin{tabular}{llll}
    \toprule
    Data-type & Model     & Eq-type & Acc. \% \\
    \midrule
     \multirow{0}{0cm}{Rot-TIM} & CNN & - & 42.5 \\
    & CNN4-aug & - & 56.7 \\
    & E2CNN & R4 & 53.5 \\
    & IEN & R4 & \textbf{56.9} \\
    & CNN8-aug & - & 58.5 \\
    & E2CNN & R8 & 56.5 \\
    & IEN & R8 & \textbf{59.7} \\
    \cmidrule{2-4}
    \multirow{0}{0cm}{R2-TIM} & CNN & - & 43.2 \\
    & CNN-aug & - & \textbf{56.3} \\
    & E2CNN & R4R & 55.9 \\
    & IEN & R4R & 56.1 \\
    \bottomrule
  \end{tabular}
% \end{minipage}\hfill
% \begin{minipage}{0.49\textwidth}
 \caption{Performance scores for VGG and its equivariant versions on Rotation (Rot-TIM) and Rotation+Reflection (R2-TIM) versions of TinyImageNet (TIM) dataset. Here, CNN4-aug and CNN8-aug denote regular CNN models similar in architecture to those of R4 and R8, respectively, at inference phase.  R4  and  R8  denote  equivariance  to  4  and  8  equidistant orientations, respectively, and R4R denotes equivariance to 4 equidistant rotations and reflections.}
  \label{table-vgg}
  \centering
  \begin{tabular}{llll}
    \toprule
    Data-type & Model     & Eq-type & Acc. \% \\
    \midrule
     \multirow{0}{0cm}{Rot-TIM} & CNN & - & 32.1   \\
    & CNN4-aug & - & 45.3 \\
    & E2CNN & R4 & 46.7 \\
    & IEN & R4 & \textbf{47.4} \\
    & CNN8-aug & - & 51.5 \\
    & E2CNN & R8 & 50.2 \\
    & IEN & R8 & \textbf{51.6} \\
    \cmidrule{2-4}
    \multirow{0}{0cm}{R2-TIM} & CNN & - & 32.5 \\
    & CNN-aug & - & 50.0 \\
    & E2CNN & R4R & \textbf{51.1} \\
    & IEN & R4R & 49.9 \\
    \bottomrule
  \end{tabular}
% \end{minipage}

\end{table}
 
% \begin{table}
%   \caption{Performance scores for ResNet-18 and its various rotation equivariant versions on rot-TIM dataset. Compositions for the architectures are described in Appendix \ref{app-resnet18}.}
%   \label{table-resnet1}
%   \centering
%   \begin{tabular}{lllcll}
%     \toprule
%     %\multicolumn{2}{c}{Part}                   \\
%     %\cmidrule(r){1-2}
%     & Data-type & Model     & Eq-type & Acc. \% \\
%     \midrule
%     \multirow{10}{*}{ResNet18} & \multirow{6}{*}{Rot} & CNN & - & -   \\
%     & & CNN-aug & - \\
%     & & E2CNN & R4 \\
%     & & E2CNN & R8 \\
%     & & EEN & R4 \\
%     & & EEN & R8 \\
%     \cmidrule{2-5}
%     & \multirow{4}{*}{RR} & CNN & - & - \\
%     & & CNN-aug & - \\
%     & & E2CNN & R4R & - \\
%     & & EEN & R4R & - \\
%     \midrule
%     \multirow{5}{*}{VGG} & E2CNN & R4 & -  &   -  \\
%     & E2CNN & R8 & - & \\
%     & EEN & R4 & - & - \\
%     & EEN & R8 & - & -\\
%     & EEN & R4-2-1 & - & - \\
%     & EEN & R8-4-2-1 & - & - \\ 
%     \bottomrule
%   \end{tabular}
%\end{table}

%% file: latex/table_track.tex
\begin{table}
\begin{minipage}{0.49\textwidth}
% {wraptable}{r}{0.5\textwidth}
% \vspace{-1em}
 \caption{Results for Object Tracking. Scores are presented for IEN implementations and two baselines: SiamFC \protect\cite{bertinetto2016fully} and RE-SiamFC \protect\cite{gupta2021rotation}}
  \label{table-track}
  \centering
  \begin{tabular}{lccc}
    \toprule
    Model & Channels\% & Succ. & Pr. \\
    \midrule
    SiamFC & - & 0.29 & 0.47 \\
    CNN4-SiamFC & 100 & 0.33 & 0.56 \\
    RE-SiamFC & 100 & 0.35 & 0.62 \\
    IEN-SiamFC & 100 & 0.37 & 0.63 \\
    Het-SiamFC & 69 & 0.37 & 0.62 \\
    \bottomrule
  \end{tabular}
  \vspace{-1em}
\end{minipage}

\end{table}

%% file: latex/table_het.tex
\begin{table}[h]
%\vspace{0em}
  \caption{Performance scores for Heterogeneous IEN implementations on Rot-TIM dataset, created through reducing the size of fraction of feature groups per layer of the network. For example, R4-2-1 implies that R4 is modified such that 50\% feature groups per layer are R4, 25\% are R2, and rest are R1, thus reducing channels per layer to 69\%.}
  \label{table-het}
  \centering
  \begin{tabular}{llcc}
    \toprule
    %\multicolumn{2}{c}{Part}                   \\
    %\cmidrule(r){1-2}
    Model & Eq-type & Channels\% & Acc.\% \\
    \midrule
    \multicolumn{4}{c}{\underline{ResNet18 variants}} \\
    E2CNN & R4 & 100  & 53.5 \\
    IEN & R4 & 100 & 56.9 \\
    Het-IEN & R4-2-1 & 69 & 55.4 \\
    \cmidrule{4-4}
    E2CNN & R8 & 100  & 56.5 \\
    IEN & R8 & 100 & 59.7 \\
    Het-IEN & R8-4-2-1 & 67 & 57.7 \\ 
    \midrule
    \multicolumn{4}{c}{\underline{VGG variants}} \\
    E2CNN & R4 & 100  &   46.7  \\
    IEN & R4 & 100 & 47.4 \\
    Het-IEN & R4-2-1 & 69 & 46.5 \\
    \cmidrule{4-4}
    E2CNN & R8 & 100 & 50.2 \\
    IEN & R8 & 100 & 51.6 \\
    Het-IEN & R8-4-2 & 69 & 51.3 \\
    Het-IEN & R8-4-2-1 & 67 & 50.8 \\ 
    \bottomrule
  \end{tabular}
  \vspace{-1em}
\end{table}

%% file: latex/table_mnist1.tex
\begin{table*}
 \caption{Performance scores of E2CNN\protect\cite{weiler2019general}, standard CNN and IEN on Rot-MNIST validation dataset along with extent of equivariance achieved in each \texttt{conv} layer as measured by the equivariance loss for that layer.}
  \label{table-mnist}
  \centering
  \begin{tabular}{lccccccc}
    \toprule
    \multirow{0}{0cm}{Model}& \multirow{0}{1cm}{Acc.\%} &\multicolumn{6}{c}{Equivariance $\mathcal{L}_G$(lower values are better)} \\
    & & \texttt{conv1} & \texttt{conv2} & \texttt{conv3} & \texttt{conv4} & \texttt{conv5} & \texttt{conv6} \\
    \midrule
    CNN (No aug) & 96.3 & $10^{0}$ & $10^{-1}$ & $10^{0}$ & $10^{0}$ & $10^{0}$ & $10^{-1}$\\
    CNN (with aug) & 98.4 & $10^{0}$ & $10^{-1}$ & $10^{0}$ & $10^{0}$ & $10^{0}$ & $10^{-1}$\\
    E2CNN  & 98.8 & $10^{-11}$ & $10^{-10}$ & $10^{-8}$ & $10^{-9}$ & $10^{-8}$ & $10^{-8}$\\
    IEN ($\beta_i$ = 1) & 98.6 & $10^{-7}$ & $10^{-6}$ & $10^{-5}$ & $10^{-5}$ & $10^{-5}$ & $10^{-4}$\\
    IEN ($\beta_i$ = 0.1) & 98.5 & $10^{-5}$ & $10^{-5}$ & $10^{-4}$ & $10^{-4}$ & $10^{-3}$ & $10^{-2}$\\
    IEN ($\beta_i$ = 0.01) & 98.6 & $10^{-3}$ & $10^{-3}$ & $10^{-2}$ & $10^{-2}$ & $10^{-2}$ & $10^{-1}$\\
    Het-IEN ($\beta_i$ = 1) & 98.6 & $10^{-6}$ & $10^{-6}$ & $10^{-5}$ & $10^{-5}$ & $10^{-5}$ & $10^{-4}$\\
    \bottomrule
  \end{tabular}
  \vspace{-1em}
\end{table*}

%% file: latex/rotMNIST_exp.tex
\begin{figure*}
    \centering
    \caption{Extended results for the MNIST feature map representations shown in the main paper. We show here the feature maps for all 6 layers of the network for the 4 orientations of the input image. }
    \begin{subfigure}{1\textwidth}
    \centering
    \includegraphics[scale=0.15]{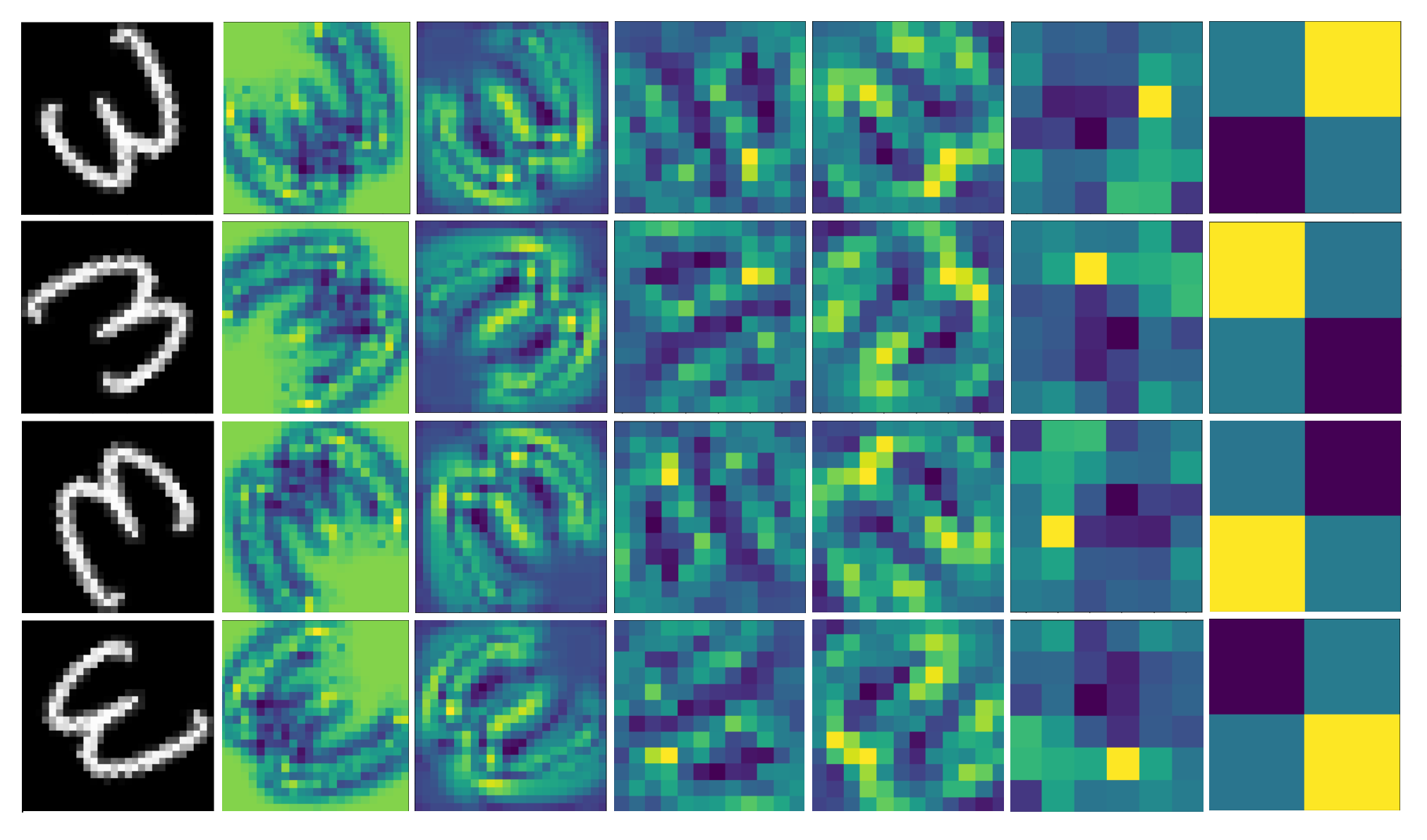}
    \caption{Steerable CNN (E2CNN \protect \cite{weiler2019general})}
    \end{subfigure}
    \begin{subfigure}{1\textwidth}
    \centering
    \includegraphics[scale=0.15]{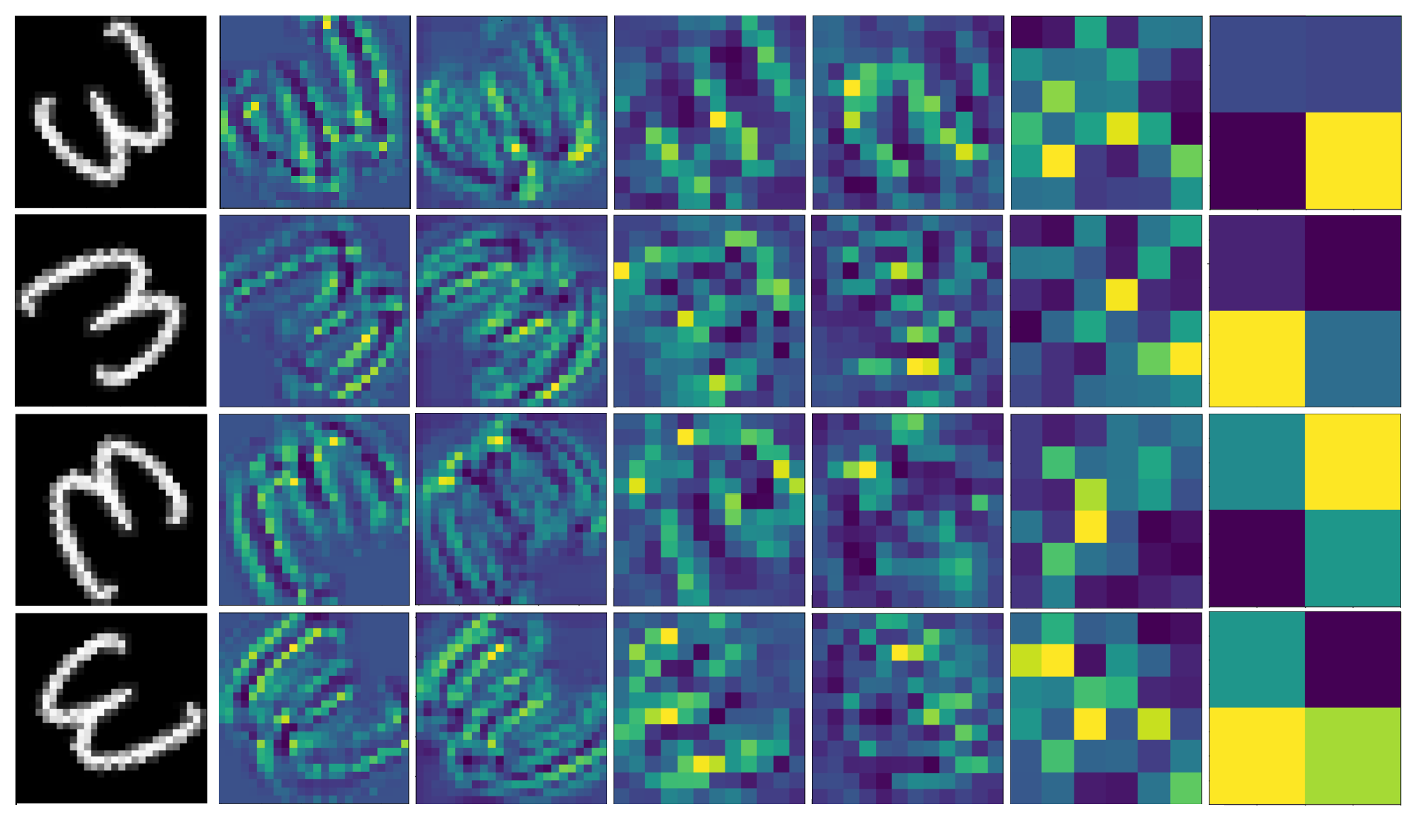}
    \caption{Regular CNN with augmentation (equivalent to IEN with $\beta_i = 0$).}
    \end{subfigure}
    \begin{subfigure}{1\textwidth}
    \centering
    \includegraphics[scale=0.15]{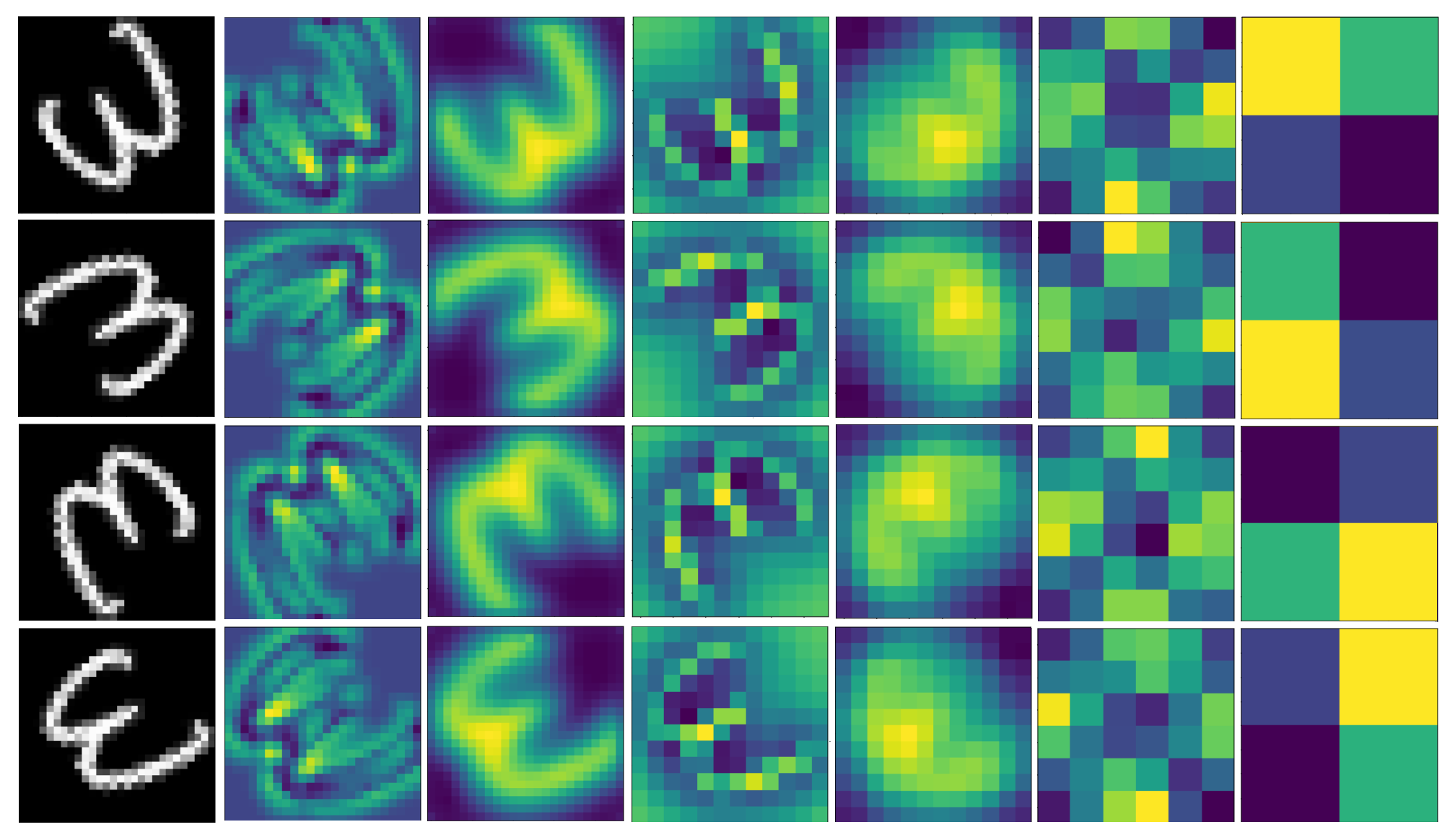}
    \caption{Implicit Equivariant Network (IEN); $\beta_i = 1.0$.}
    \end{subfigure}
    
    \label{fig-mnist-equi}
\end{figure*}

%% file: latex/Resnet18_exp.tex
\begin{figure*}
    \centering
    \caption{Feature maps outputs from 4 blocks of E2CNN variant of ResNet18 shown for 8 orientations of an input.}
    \includegraphics[scale=0.43]{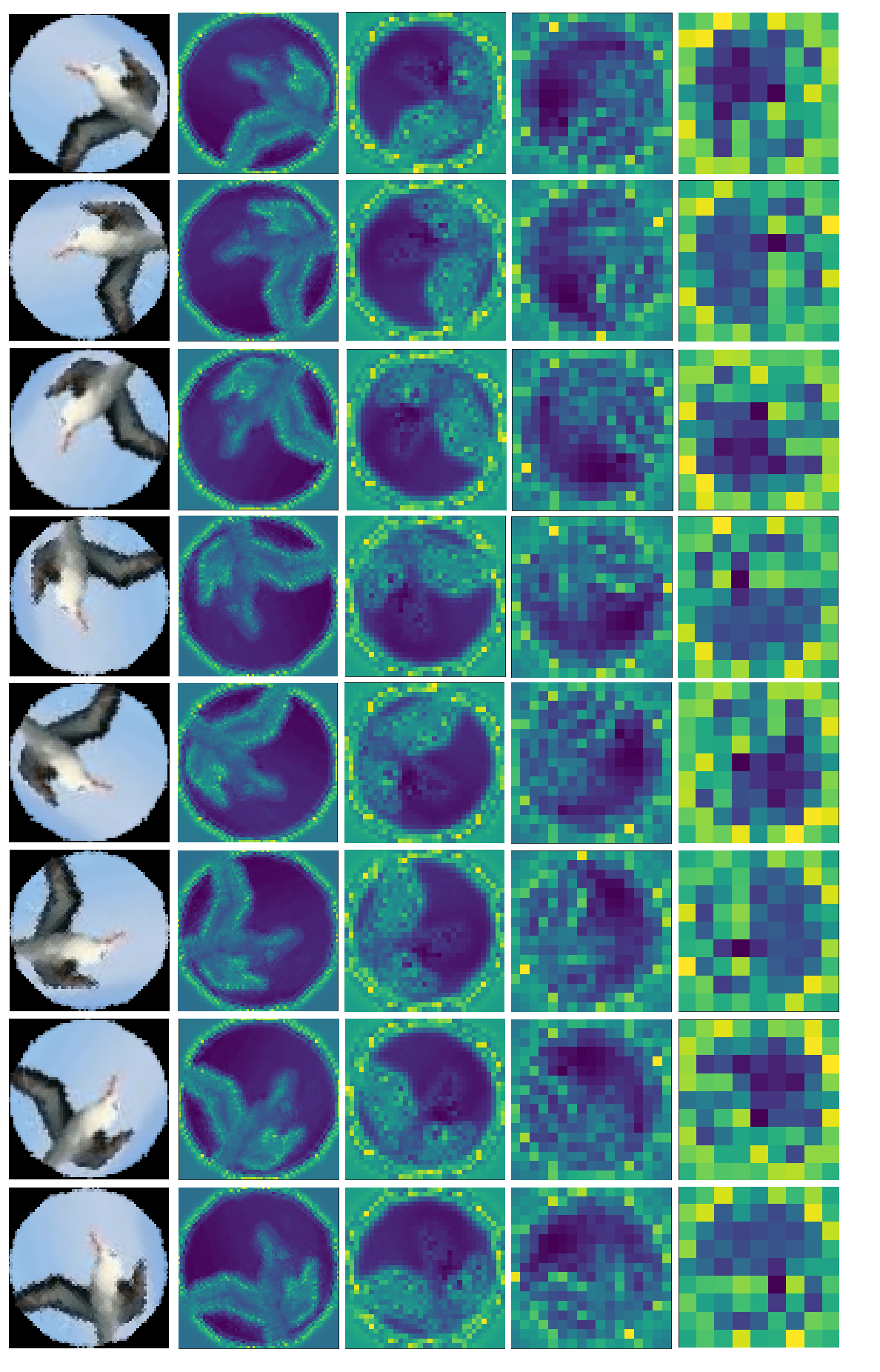}
    
    \label{fig-resnet-e2cnn}
\end{figure*}
\begin{figure*}
    \centering
    \caption{Feature maps outputs from 4 blocks of regular CNN-8 data augmentation version of ResNet18 shown for 8 orientations of an input.}
    \includegraphics[scale=0.43]{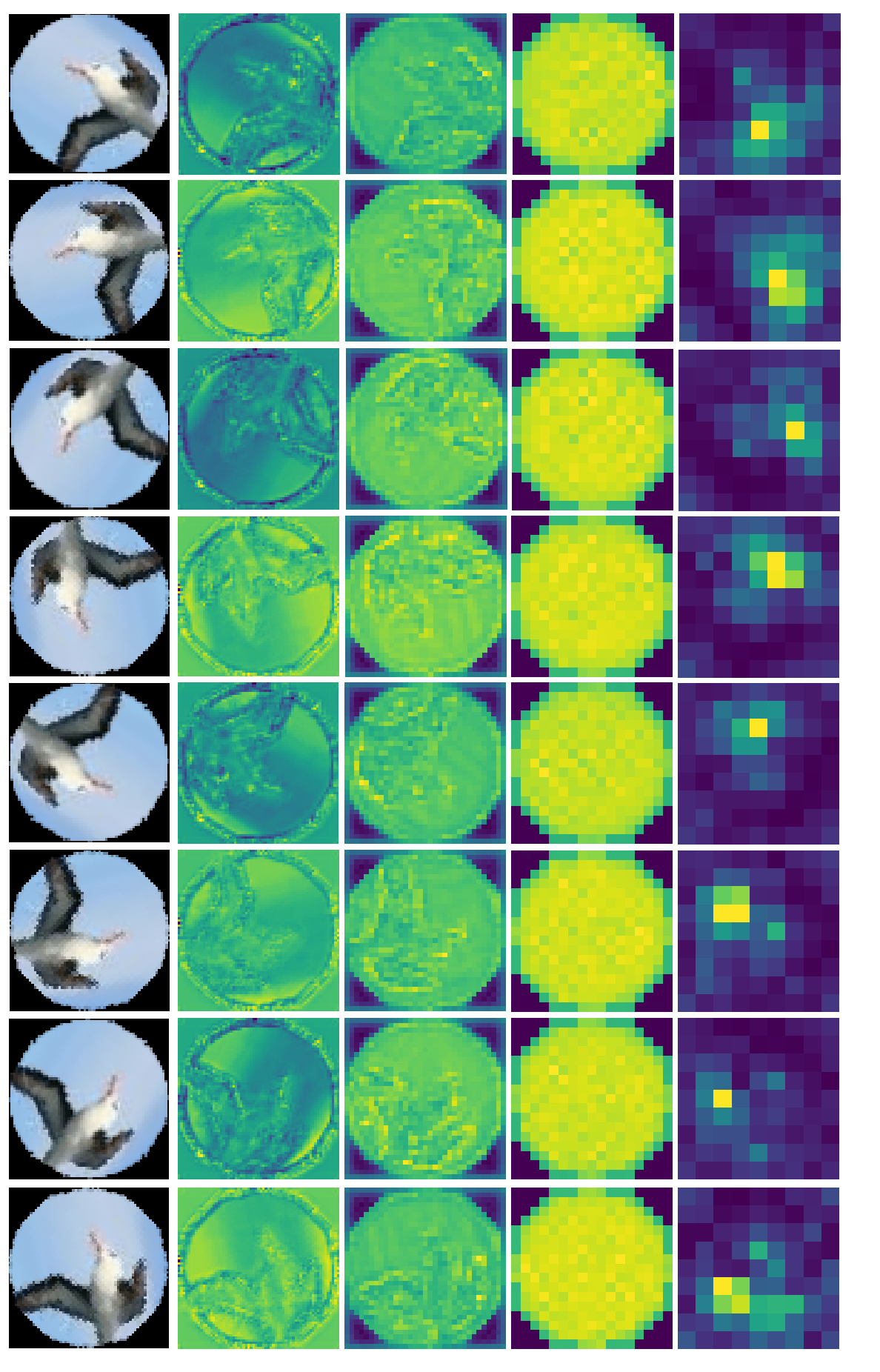}
    
    \label{fig-resnet-cnn8}
\end{figure*}
\begin{figure*}
    \centering
    \caption{Feature maps outputs from 4 blocks of implicitly equivariant (IEN-R8) version of ResNet18 shown for 8 orientations of an input (all $\beta_i = 0.01$.)}
    \includegraphics[scale=0.43]{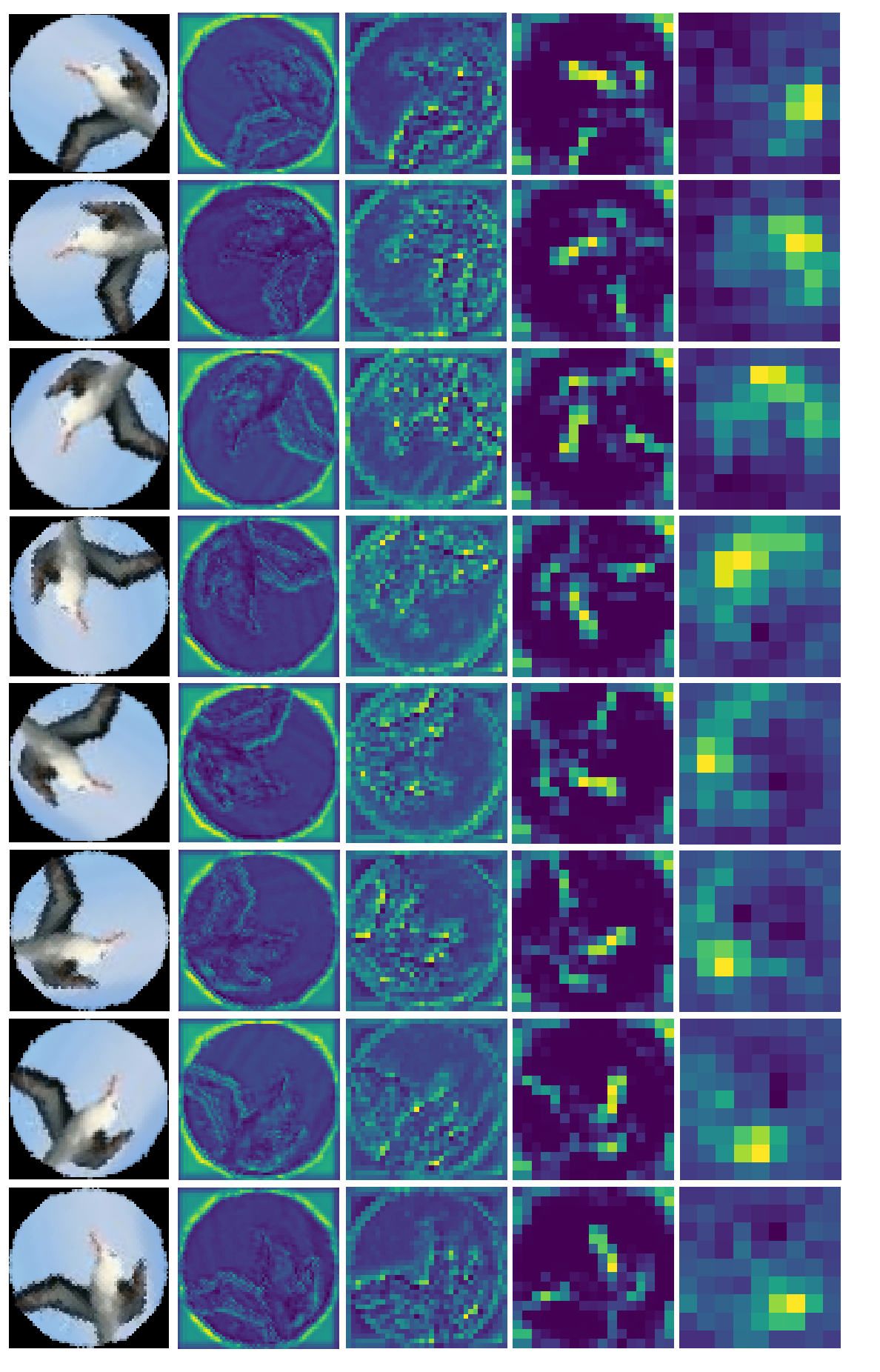}
    
    \label{fig-resnet-een001}
\end{figure*}
\begin{figure*}
    \centering
    \caption{Feature maps outputs from 4 blocks of implicitly equivariant (IEN-R8) version of ResNet18 shown for 8 orientations of an input (all $\beta_i = 0.1$.)}
    \includegraphics[scale=0.43]{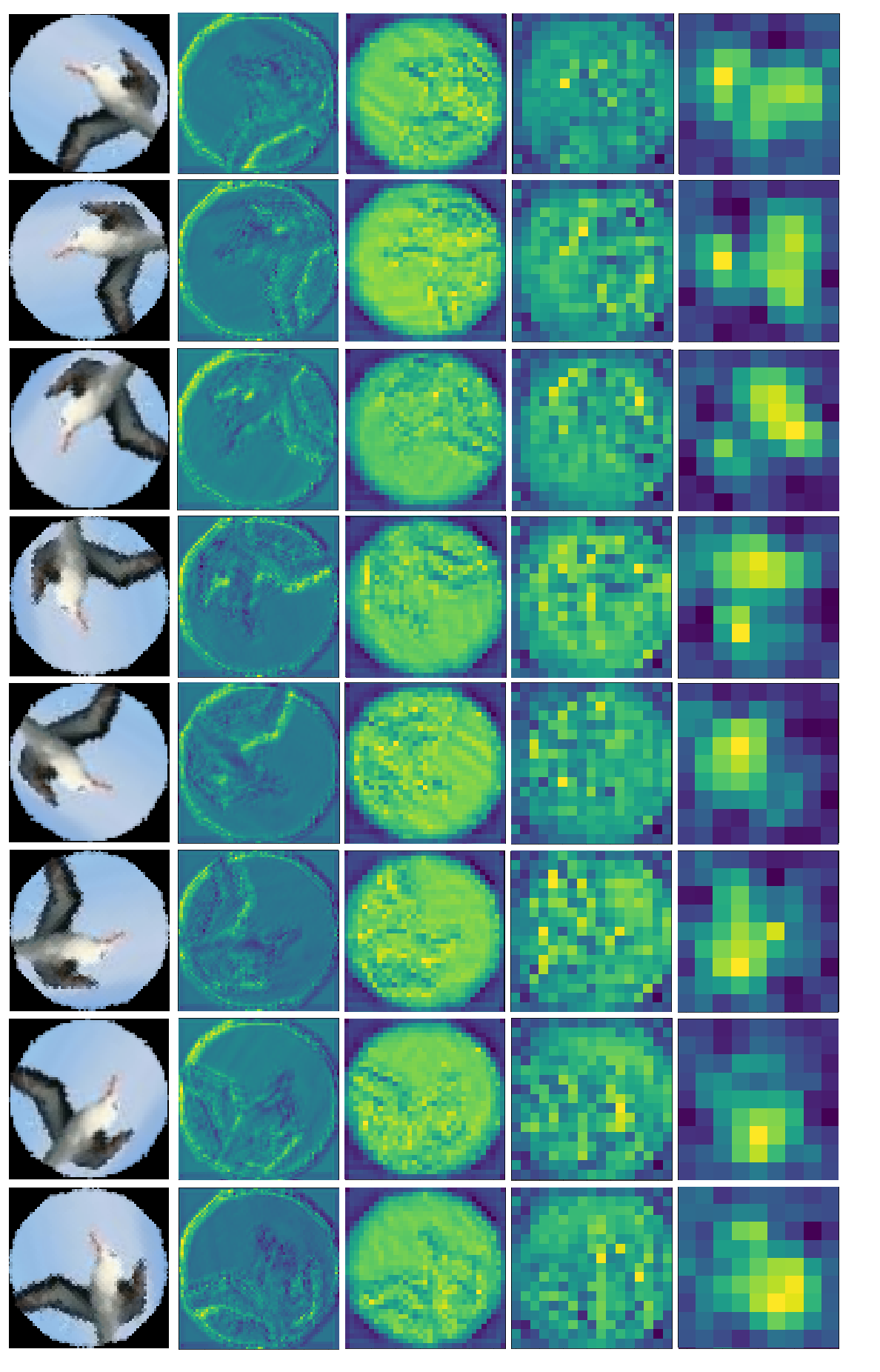}
    
    \label{fig-resnet-een01}
\end{figure*}
\begin{figure*}
    \centering
    \caption{Feature maps outputs from 4 blocks of implicitly equivariant (IEN-R8) version of ResNet18 shown for 8 orientations of an input (all $\beta_i = 1.0$.)}
    \includegraphics[scale=0.43]{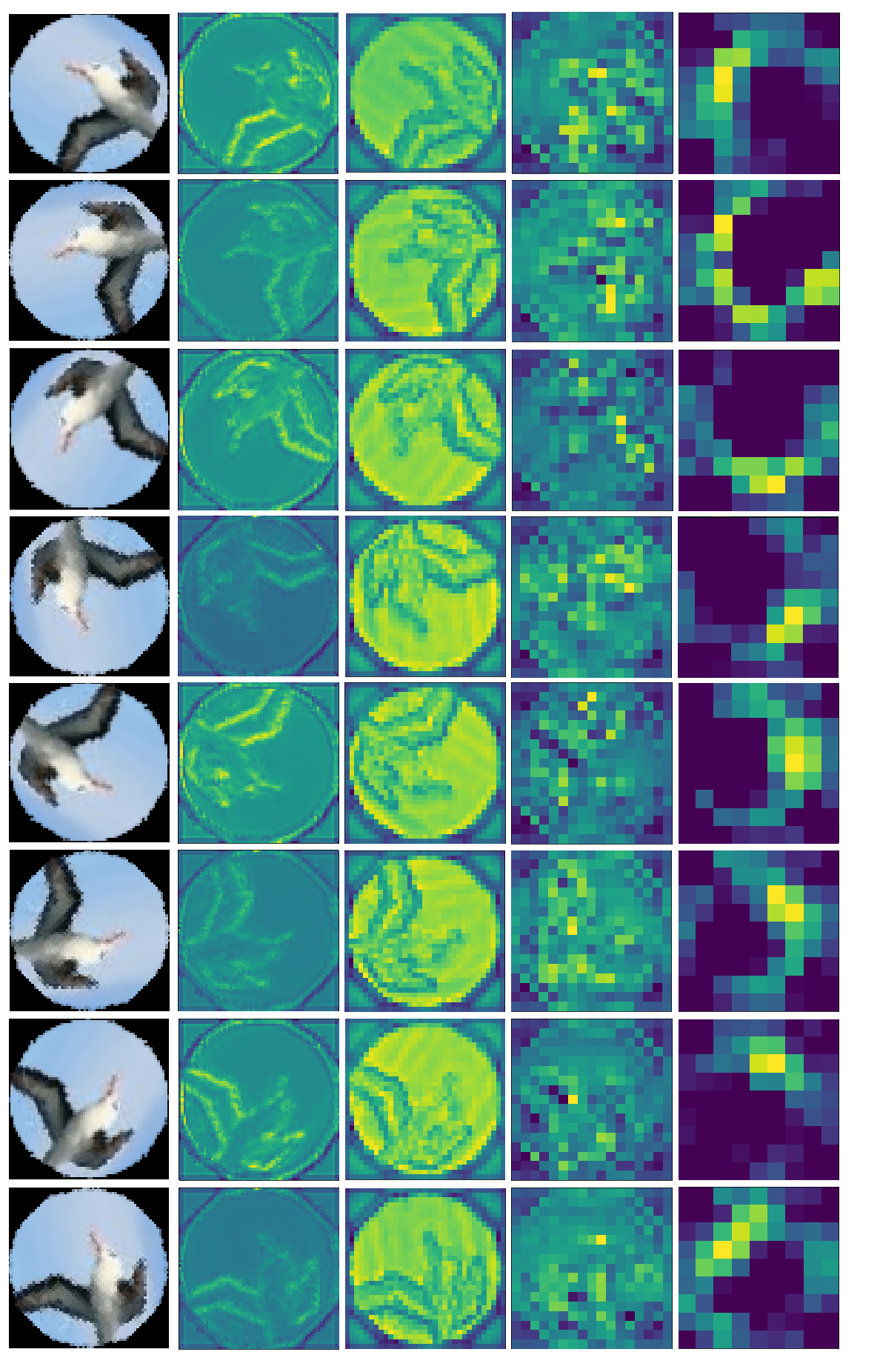}
    
    \label{fig-resnet-een1}
\end{figure*}

An E2CNN model equivalent to ResNet18 and VGG was developed with the same layers and equal number of total trainable parameters. To maintain equal parameters count, a ratio was found in channels of the standard architecture and feature fields of E2CNN for different equivariance type. Full details about number of channels per layer in the base model and in E2CNN corresponding to equal trainable parameter count is provided in Table \ref{table-resnet-params} and \ref{table-vgg-params}.
In addition, we developed the IEN architecture such that it exactly matches the architecture of E2CNN at inference time. %Full results for the experiments on Rot-TIM and RR-TIM dataset for E2CNN, standard CNN and IEN experiments with different values of $\beta_i$ and the extent of equivariance achieved in all \texttt{conv} blocks in case of ResNet18 and in 4 alternate conv layers in case of VGG (\texttt{conv}1, \texttt{conv}3, \texttt{conv}5, \texttt{conv}7) are reported in Table \ref{table-rottim}.
Full results for the experiments on Rot-TIM and RR-TIM dataset for E2CNN, standard CNN and IEN experiments with different values of $\beta_i$ are shown in Table \ref{table-rottim}. The extent of equivariance achieved in all \texttt{conv} blocks in case of ResNet18 and in 4 alternate conv layers in case of VGG (\texttt{conv}1, \texttt{conv}3, \texttt{conv}5, \texttt{conv}7) are also reported in Table \ref{table-rottim}.
Since the extent of equivariance is computed using the equivariance loss, lower values are better. The extent of equivariance observed in case of R8 in ResNet18 based E2CNN and standard CNN models is shown in Figure \ref{fig-resnet-e2cnn} and Figure \ref{fig-resnet-cnn8}, respectively. The corresponding results for IEN ($\beta_i$ = 0.01), IEN ($\beta_i$ = 0.1) are provided in Figure \ref{fig-resnet-een001}, Figure \ref{fig-resnet-een01} and Figure \ref{fig-resnet-een1}, respectively.
The extent of equivariance observed in case of R8 in ResNet18 based E2CNN, standard CNN, IEN ($\beta_i$ = 0.01), IEN ($\beta_i$ = 0.1), IEN ($\beta_i$ = 1) for each \texttt{conv} block is shown in Fig \ref{fig-resnet-e2cnn}, Fig \ref{fig-resnet-cnn8}, Fig \ref{fig-resnet-een001}, Fig \ref{fig-resnet-een01} and Fig \ref{fig-resnet-een1} respectively. \\

%% file: latex/table_resnet_params.tex
\begin{table*}
%\vspace{0em}
  \caption{Details on the composition of E2CNN architectures based on ResNet18 model experimented in this paper. Here R4 and R8 denote equivariance to 4 and 8 equidistant orientations, respectively, and R4R denotes equivariance to 4 equidistant rotations and reflections. Numbers for \texttt{conv} layers denote channels per orientation. Channels per layer in all variant are chosen such that the total parameters in the model are same as the base ResNet18 model}
  \label{model-arch}
  \centering
  \begin{tabular}{lccccc}
    \toprule
    %\multicolumn{2}{c}{Part}                   \\
    %\cmidrule(r){1-2}
    Layer & kernel size & Resnet-18 & E2CNN R4 & E2CNN R8 & E2CNN R4R \\
    \midrule
    \texttt{conv-1} & 3$\times$3 & 64 & 34 & 24 & 24\\
    \midrule
    \multirow{0}{0cm}{\texttt{block-1}} & 3$\times$3 & 64 & 34 & 24 & 24\\
    & 3$\times$3 & 64 & 34 & 24 & 24\\
    \midrule
    \multirow{0}{0cm}{\texttt{block-2}} & 3$\times$3 & 128 & 68 & 48 & 48\\
    & 3$\times$3 & 128 & 68 & 48 & 48\\
    \midrule
    \multirow{0}{0cm}{\texttt{block-3}} & 3$\times$3 & 256 & 136 & 96 & 96\\
    & 3$\times$3 & 256 & 136 & 96 & 96\\
    \midrule
    \multirow{0}{0cm}{\texttt{block-4}} & 3$\times$3 & 512 & 272 & 192 & 192 \\
    & 3$\times$3 & 512 & 272 & 192 & 192 \\
    \midrule
    \texttt{avg-pool} & - & 512 & 272 & 192 & 192 \\
    \midrule
    \texttt{fc-layer} & - & 100 & 100 & 100 & 100 \\
    \bottomrule
  \end{tabular}
  \label{table-resnet-params}
  \vspace{4em}
\end{table*}

%% file: latex/table_vgg_params.tex
\begin{table*}
%\vspace{0em}
  \caption{Details on the composition of E2CNN architectures based on VGG model experimented in this paper. Here R4 and R8 denote equivariance to 4 and 8 equidistant orientations, respectively, and R4R denotes equivariance to 4 equidistant rotations and reflections. Numbers for \texttt{conv} layers denote channels per orientation. Channels per layer in all variant are chosen such that the total parameters in the model are same as the base VGG model.}
  \label{model-arch}
  \centering
  \begin{tabular}{lccccc}
    \toprule
    %\multicolumn{2}{c}{Part}                   \\
    %\cmidrule(r){1-2}
    Layer & kernel size & VGG & E2CNN R4 & E2CNN R8 & E2CNN R4R \\
    \midrule
    \texttt{conv-1} & 3$\times$3 & 64 & 36 & 25 & 25\\
    \texttt{max-pool} & 2$\times$2 & 64 & 36 & 25 & 25\\
    \midrule
    \texttt{conv-2} & 3$\times$3 & 128 & 72 & 50 & 50\\
    \texttt{max-pool} & 2$\times$2 & 128 & 72 & 50 & 50\\
    \midrule
    \texttt{conv-3} & 3$\times$3 & 256 & 144 & 100 & 100\\
    \texttt{conv-4} & 3$\times$3 & 256 & 144 & 100 & 100\\
    \texttt{max-pool} & 2$\times$2 & 256 & 144 & 100 & 100\\
    \midrule
    \texttt{conv-5} & 3$\times$3 & 512 & 288 & 200 & 200\\
    \texttt{conv-6} & 3$\times$3 & 512 & 288 & 200 & 200\\
    \texttt{max-pool} & 2$\times$2 & 512 & 288 & 200 & 200\\
    \midrule
    \texttt{conv-7} & 3$\times$3 & 512 & 288 & 200 & 200\\
    \texttt{conv-8} & 3$\times$3 & 512 & 288 & 200 & 200\\
    \texttt{avg-pool} & - & 512 & 288 & 200 & 200\\
    \midrule
    \texttt{fc-layer} & - & 1024 & 1024 & 1024 & 1024\\
    \midrule
    \texttt{fc-layer} & - & 100 & 100 & 100 & 100\\
    % \multirow{2}{*}{\texttt{block-1}} & 3$\times$3 & 64 & 34 & 24 & 24\\
    % & 3$\times$3 & 64 & 34 & 24 & 24\\
    % \midrule
    % \multirow{2}{*}{\texttt{block-2}} & 3$\times$3 & 128 & 68 & 48 & 48\\
    % & 3$\times$3 & 128 & 68 & 48 & 48\\
    % \midrule
    % \multirow{2}{*}{\texttt{block-3}} & 3$\times$3 & 256 & 136 & 96 & 96\\
    % & 3$\times$3 & 256 & 136 & 96 & 96\\
    % \midrule
    % \multirow{2}{*}{\texttt{block-4}} & 3$\times$3 & 512 & 272 & 192 & 192 \\
    % & 3$\times$3 & 512 & 272 & 192 & 192 \\
    % \midrule
    % \texttt{avg-pool} & - & 512 & 272 & 192 & 192 \\
    % \midrule
    % \texttt{fc-layer} & - & 100 & 100 & 100 & 100 \\
    \bottomrule
  \end{tabular}
  \label{table-vgg-params}
   \vspace{2.5em}
\end{table*}

%% file: latex/sec_app_scale_var2.tex
\begin{table*}[h]
 \caption{Classification error on Scale-MNIST and STL-10 datasets for our IEN and three baseline methods, namely SS-CNN \protect\cite{ghosh2019arxiv}, DSS \protect\cite{worrall2019neurips} and SESN \protect\cite{Sosnovik2020Scale-Equivariant}. For Scale-MNIST, we use two variants: image sizes of $28 \times 28$ and $56 \times 56$. All baseline implementations are based on the description provided in \protect\cite{Sosnovik2020Scale-Equivariant}.}
  \label{table-scale}
  \centering
  \begin{tabular}{llc}
    \toprule
    Dataset & Model & Succ. \\
    \midrule
    \multirow{4}{*}{Scale-MNIST ($28 \times 28$)} & SS-CNN & 2.10 \\
    & DSS & 1.95 \\
    & SESN & 1.76 \\
    & IEN & 1.78\\
    \midrule
    \multirow{4}{*}{Scale-MNIST ($56 \times 56$)} & SS-CNN & 1.76 \\
    & DSS & 1.57 \\
    & SESN & 1.42 \\
    & IEN & 1.33 \\
    \midrule
    \multirow{4}{*}{STL-10} & SS-CNN & 25.47 \\
    & DSS & 11.28\\
    & SESN & 10.83 \\
    & IEN & 10.11 \\
    \bottomrule
  \end{tabular}
\end{table*}

%% file: latex/table_rottim_appendix2.tex
\begin{table*}
%\begin{minipage}{0.49\textwidth}
 \caption{Performance scores and equivariance loss achieved for ResNet18 and its equivariant versions as well as VGG and its equivariant versions on Rotation (Rot-TIM) and Rotation+Reflection (R2-TIM)  are shown. Here R4 and R8 denote equivariance to 4 and 8 equidistant rotations and R4R denotes equivariance to 4 equidistant rotations and reflections. The extent of equivariance achieved in the 4 conv blocks of ResNet18 and in 4 alternate conv layers in case of VGG are reported for each model, lower values are better.}
  \label{table-rottim}
  \centering
  \begin{tabular}{clcccc}
    \toprule
    % \multicolumn{4}{c}{}
    \multirow{2}{*}{Eq-type} & \multirow{2}{*}{Model} & \multicolumn{2}{c}{ResNet18} & \multicolumn{2}{c}{VGG} \\
    & & Acc \% & Equi.($\mathcal{L}_G$) & Acc \% & Equi. \\
    \midrule
    \multirow{6}{*}{R4} & CNN (No Aug) & 42.5 & $10^3$,$10^3$,$10^3$,$10^0$ & 32.1 & $10^1$,$10^2$,$10^2$,$10^1$  \\
    & CNN (with aug) & 56.7 & $10^3$,$10^3$,$10^3$,$10^0$ & 45.3 & $10^1$,$10^2$,$10^2$,$10^1$ \\
    & E2CNN & 53.5 & $10^{-11}$,$10^{-10}$,$10^{-9}$,$10^{-8}$ & 46.7 & $10^{-11}$,$10^{-9}$,$10^{-9}$,$10^{-8}$  \\
    & IEN ($\beta_i$ = 0.01) & 56.9 & $10^{-3}$,$10^{-4}$,$10^{-3}$,$10^{-1}$ & 45.8 & $10^{-2}$,$10^{-2}$,$10^{-2}$,$10^{-1}$ \\
    & IEN ($\beta_i$ = 0.1) & 55.8 & $10^{-4}$,$10^{-5}$,$10^{-5}$,$10^{-2}$ & 47.4 & $10^{-4}$,$10^{-3}$,$10^{-3}$,$10^{-2}$ \\
    & IEN ($\beta_i$ = 1) & 54.7 & $10^{-5}$,$10^{-6}$,$10^{-5}$,$10^{-3}$ & 46.1 & $10^{-5}$,$10^{-4}$,$10^{-4}$,$10^{-3}$ \\
    \\
    \multirow{6}{*}{R8} & CNN (No Aug) & 42.6 & $10^3$,$10^3$,$10^3$,$10^0$ & 32.1 & $10^1$,$10^2$,$10^2$,$10^1$  \\
    & CNN (with aug) & 58.5 & $10^3$,$10^3$,$10^3$,$10^0$ & 51.5 & $10^1$,$10^2$,$10^2$,$10^1$ \\
    & E2CNN & 56.5 &  $10^{-10}$,$10^{-10}$,$10^{-9}$,$10^{-9}$ & 50.2 & $10^{-11}$,$10^{-9}$,$10^{-9}$,$10^{-8}$ \\
    & IEN ($\beta_i$  = 0.01) & 58.6 & $10^{-3}$,$10^{-4}$,$10^{-3}$,$10^{-1}$ & 51.6 & $10^{-2}$,$10^{-2}$,$10^{-1}$,$10^{-1}$ \\
    & IEN ($\beta_i$ = 0.1) & 59.7 & $10^{-5}$,$10^{-5}$,$10^{-4}$,$10^{-2}$ & 51.1 & $10^{-3}$,$10^{-3}$,$10^{-2}$,$10^{-2}$ \\
    & IEN ($\beta_i$ = 1) & 58.6 & $10^{-5}$,$10^{-6}$,$10^{-5}$,$10^{-2}$ & 48.9 & $10^{-4}$,$10^{-4}$,$10^{-3}$,$10^{-3}$ \\
    \\
    \multirow{6}{*}{R4R} & CNN (No Aug) & 43.2 & $10^3$,$10^3$,$10^3$,$10^0$ & 32.5 & $10^1$,$10^2$,$10^2$,$10^1$  \\
    & CNN (with aug) & 56.3 & $10^3$,$10^3$,$10^3$,$10^0$ & 50.0 & $10^1$,$10^2$,$10^2$,$10^1$ \\
    & E2CNN & 55.9 & $10^{-11}$,$10^{-10}$,$10^{-9}$,$10^{-9}$ & 51.1 & $10^{-11}$,$10^{-9}$,$10^{-9}$,$10^{-9}$ \\
    & IEN ($\beta_i$ = 0.01) & 56.1 & $10^{-3}$,$10^{-4}$,$10^{-3}$,$10^{-1}$ & 49.9 & $10^{-2}$,$10^{-2}$,$10^{-1}$,$10^{-1}$ \\
    & IEN ($\beta_i$ = 0.1) & 55.7 & $10^{-4}$,$10^{-5}$,$10^{-4}$,$10^{-2}$ & 49.7 & $10^{-3}$,$10^{-3}$,$10^{-2}$,$10^{-1}$ \\
    & IEN ($\beta_i$ = 1) & 54.0 & $10^{-5}$,$10^{-6}$,$10^{-5}$,$10^{-3}$ & 48.3 & $10^{-5}$,$10^{-4}$,$10^{-3}$,$10^{-3}$ \\
    % &&&&&&&\\
    % \multirow{2}{*}{R8} & ResNet18 & 56.5 & 42.9 & 58.5 & 58.6 & 59.7 & 58.5 \\
    % & VGG & 53.5 & 42.5 & 56.7 & 56.9 & 55.6 &  \\
    % &&&&&&&\\
    % \multirow{2}{*}{R4R} & ResNet18 & 55.9 & 43.2 & 56.1 & 55.7 & 52.0 &  \\
    % & VGG & 53.5 & 42.5 & 56.7 & 56.9 & 55.6 &  \\
    % % & CNN4-aug & - & 56.7 \\
    % % & E2CNN & R4 & 53.5 \\
    % % & EEN & R4 & \textbf{56.9} \\
    % % & CNN8-aug & - & 58.5 \\
    % % & E2CNN & R8 & 56.5 \\
    % % & EEN & R8 & \textbf{59.7} \\
    % % \cmidrule{2-9}
    % % \multirow{4}{*}{R2-TIM} & CNN & - & 43.2 \\
    % % & CNN-aug & - & \textbf{56.3} \\
    % % & E2CNN & R4R & 55.9 \\
    % % & EEN & R4R & 56.1 \\
    \bottomrule
  \end{tabular}
% \end{minipage}\hfill
% \begin{minipage}{0.49\textwidth}
%  \caption{Performance scores for VGG and its equivariant versions on Rotation (Rot-TIM) and Rotation+Reflection (R2-TIM) versions of TinyImageNet (TIM) dataset. Here, CNN4-aug and CNN8-aug denote regular CNN models similar in architecture to those of R4 and R8, respectively, at inference phase.}
%   \label{table-vgg}
%   \centering
%   \begin{tabular}{llll}
%     \toprule
%     Data-type & Model     & Eq-type & Acc. \% \\
%     \midrule
%      \multirow{6}{*}{Rot-TIM} & CNN & - & 32.1   \\
%     & CNN4-aug & - & 45.3 \\
%     & E2CNN & R4 & 46.7 \\
%     & EEN & R4 & \textbf{47.4} \\
%     & CNN8-aug & - & 51.5 \\
%     & E2CNN & R8 & 50.2 \\
%     & EEN & R8 & \textbf{51.6} \\
%     \cmidrule{2-4}
%     \multirow{4}{*}{R2-TIM} & CNN & - & 32.5 \\
%     & CNN-aug & - & 50.0 \\
%     & E2CNN & R4R & \textbf{51.1} \\
%     & EEN & R4R & 49.9 \\
%     \bottomrule
%   \end{tabular}
% \end{minipage}

\end{table*}